\documentclass[10pt,twocolumn,letterpaper]{article}

\usepackage[pagenumbers]{cvpr} 
\newcommand{\ours}{MagicArticulate}
\newcommand{\ourdata}{Articulation-XL}
\newcommand{\res}{ModelsResource}
\usepackage[pagebackref=true,breaklinks=true,letterpaper=true,colorlinks,bookmarks=false]{hyperref}
\usepackage{times}
\usepackage{epsfig}
\usepackage{graphicx}
\usepackage{amsmath}
\usepackage{amssymb}
\usepackage{booktabs}
\usepackage{multicol}
\usepackage{multirow}
\usepackage{caption}
\usepackage{subcaption}
\usepackage{listings}

\newcommand{\boldstartspace}[1]{\medskip\noindent\textbf{#1}}

\title{MagicArticulate: Make Your 3D Models Articulation-Ready}

\author{
    Chaoyue Song\textsuperscript{1,2}, 
    Jianfeng Zhang$^{\dag}$\textsuperscript{2}, 
    Xiu Li\textsuperscript{2}, 
    Fan Yang\textsuperscript{1}, 
    Yiwen Chen\textsuperscript{1}, 
    Zhongcong Xu\textsuperscript{2}, \\
    Jun Hao Liew\textsuperscript{2},
    Xiaoyang Guo\textsuperscript{2}, 
    Fayao Liu\textsuperscript{3}, 
    Jiashi Feng\textsuperscript{2}, 
    Guosheng Lin$^{\dag}$\textsuperscript{1} \\
    \textsuperscript{1}{Nanyang Technological University} \quad
    \textsuperscript{2}{ByteDance Seed} \\ 
    \textsuperscript{3}{Institute for Inforcomm Research, A*STAR}
}

\begin{document}

\maketitle

\let\oldthefootnote\thefootnote
\let\thefootnote\relax
\footnotetext{$^{\dag}$ Corresponding authors.}
\let\thefootnote\oldthefootnote

\begin{abstract}

With the explosive growth of 3D content creation, there is an increasing demand for automatically converting static 3D models into articulation-ready versions that support realistic animation. Traditional approaches rely heavily on manual annotation, which is both time-consuming and labor-intensive. Moreover, the lack of large-scale benchmarks has hindered the development of learning-based solutions. In this work, we present MagicArticulate, an effective framework that automatically transforms static 3D models into articulation-ready assets. Our key contributions are threefold. First, we introduce Articulation-XL, a large-scale benchmark containing over 33k 3D models with high-quality articulation annotations, carefully curated from Objaverse-XL. Second, we propose a novel skeleton generation method that formulates the task as a sequence modeling problem, leveraging an auto-regressive transformer to naturally handle varying numbers of bones or joints within skeletons and their inherent dependencies across different 3D models. Third, we predict skinning weights using a functional diffusion process that incorporates volumetric geodesic distance priors between vertices and joints. Extensive experiments demonstrate that MagicArticulate significantly outperforms existing methods across diverse object categories, achieving high-quality articulation that enables realistic animation. Project page: \url{https://chaoyuesong.github.io/MagicArticulate}.

\end{abstract}    
\section{Introduction}
\label{sec:intro}

\begin{figure*}
    \centering
\includegraphics[width=\textwidth]
    {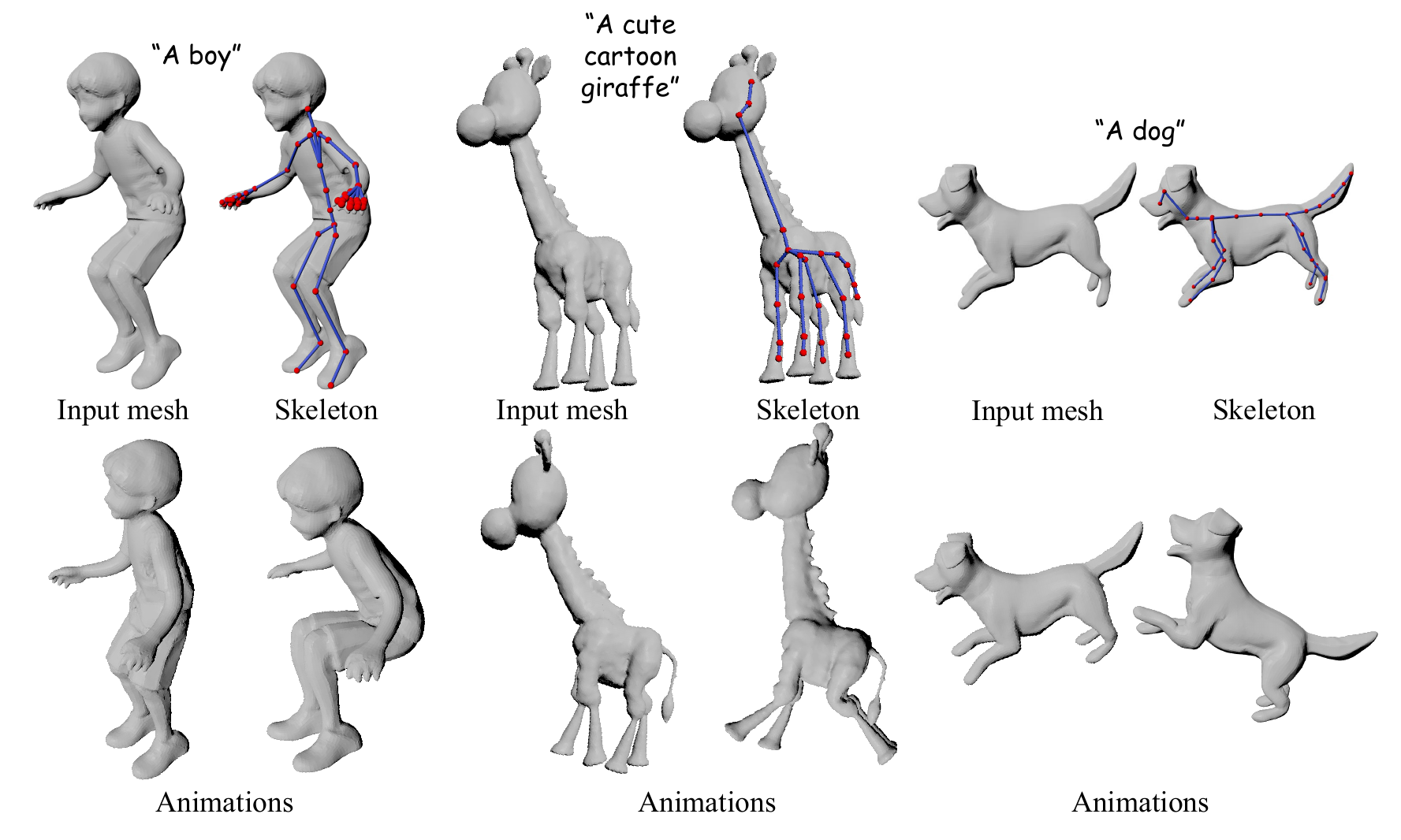}
    \caption{\textbf{Given a 3D model, \ours{} can automatically generate the skeleton and skinning weights, making the model articulation-ready without further manual refinement.} The input meshes are generated by Rodin Gen-1 \cite{zhang2024clay} and Tripo 2.0 \cite{tripo3d}. The meshes and skeletons are rendered using Maya Software Renderer \cite{AutodeskMaya2024}.}
    \label{teaser}
  \end{figure*}

The rapid advancement of 3D content creation has led to an increasing demand for articulation-ready 3D models, especially in gaming, VR/AR, and robotics simulation. Converting static 3D models into articulation-ready versions traditionally requires professional artists to manually place skeletons, define joint hierarchies and specify skinning weights, which is both time-consuming and demands significant expertise, making it a major bottleneck in modern content creation pipelines.

To address these issues, various automatic approaches for skeleton extraction have been proposed, which can be categorized into template-based \cite{baran2007automatic, li2021learning} and template-free methods \cite{xu2020rignet, xu2019predicting, huang2013l1, au2008skeleton}. Template-based methods, like Pinocchio \cite{baran2007automatic}, fit predefined skeletal templates to input shapes. While they achieve satisfactory results for specific categories like human characters, they struggle to generalize to objects with varying structural patterns. Moreover, these methods mostly rely on distance metrics between joints and vertices for skinning weight prediction, which often fail on shapes with complex topology. Many template-free methods \cite{huang2013l1, au2008skeleton, cao2010point, lin2021point2skeleton, tagliasacchi2012mean} extract curve skeletons from meshes or point clouds using shape medial axis or the centerline of shapes, but often produce densely packed joints that are unsuitable for animation.
Recent deep learning methods like RigNet \cite{xu2020rignet} have shown promise in predicting skeletons and skinning weights directly from input shapes. However, they rely heavily on carefully crafted features and make strong assumptions about shape orientation, limiting their ability to handle diverse object categories. These limitations stem from two fundamental challenges: the lack of a large-scale, diverse dataset for training generalizable models, and the inherent difficulty in designing an effective framework capable of handling complex mesh topologies, accommodating varying skeleton structures, and ensuring the coherent generation of both accurate skeletons and skinning weights.

To overcome these challenges, we first introduce Articulation-XL, a large-scale dataset containing over 33k 3D models with high-quality articulation annotations carefully curated from Objaverse-XL \cite{deitke2023objaverse, deitke2024objaverse}. Built upon this benchmark, we propose MagicArticulate, a novel framework that addresses both skeleton generation and skinning weight prediction. Specifically, we reformulate skeleton generation as an auto-regressive sequence modeling task, enabling our model to naturally handle varying numbers of bones or joints within skeletons across different 3D models. For skinning weight prediction, we develop a functional diffusion framework that learns to generate smoothly transitioning skinning weights over mesh surfaces by incorporating volumetric geodesic distance priors between vertices and joints, effectively handling complex mesh topologies that challenge traditional geometric-based methods. These designs demonstrate superior scalability on large-scale datasets and generalize well across diverse object categories, without requiring assumptions about shape orientation or topology.

Extensive experiments on our Articulation-XL and \res{} \cite{ModelsResource2019} collected by Xu et al. \cite{xu2019predicting, xu2020rignet}
demonstrate the effectiveness of MagicArticulate in both skeleton generation and skinning weight prediction. The proposed methods also generalize well to 3D models from various sources, including artist-created assets, and models generated by AI techniques. With the generated skeleton and skinning weights, our method automatically creates ready-to-animate assets that support natural pose manipulation without manual refinement (\Cref{teaser}), particularly beneficial for large-scale animation content creation. 

Our key contributions include: (1) The first large-scale articulation benchmark containing over 33k models with high-quality articulation annotations; (2) A novel two-stage framework that effectively handles both skeleton generation and skinning weight prediction; (3) State-of-the-art performance and demonstrated practicality in real-world animation pipelines.

\section{Related works}
\label{sec:related}
\subsection{Skeleton generation}
There are two categories of methods for creating skeletons in 3D models. The first category relies on predefined templates \cite{baran2007automatic, li2021learning} or additional annotations \cite{xu2022morig, de2008automatic, mixamo, james2005skinning}. Pinocchio \cite{baran2007automatic} is a pioneering method for automatically extracting an animation skeleton from an input 3D model. It fits a predefined skeleton template to the 3D model, evaluating the fitting cost for different templates and selecting the most suitable one for a given model. Li et al. \cite{li2021learning} proposed a deep learning-based method to estimate joint positions for a given human skeletal template. However, these template-based methods are limited to rigging characters whose articulation structures are compatible with the predefined templates, making it difficult to generalize to objects with distinct structures. 

There are also methods that rely on additional inputs or annotations to generate skeletons for 3D models, including point cloud sequences \cite{xu2022morig}, mesh sequences \cite{de2008automatic, james2005skinning}, and manual annotations \cite{mixamo}. Additionally, recent works \cite{yang2022banmo, song2024reacto, song2024moda, zhang2024s3o, zhang2024learning, zhang2024magicpose4d} have focused on learning the joints and bones of articulated objects directly from videos to reconstruct object motion. In contrast, our approach aims to generate skeletons using only 3D models as input.

The second category consists of template-free methods that operate without relying on predefined templates or additional annotations. Many approaches \cite{au2008skeleton, cao2010point, huang2013l1, tagliasacchi2012mean, lin2021point2skeleton} are designed to extract curve skeletons from meshes or point clouds by utilizing the medial axis or the centerline of shapes. These methods often result in densely packed joints that are unsuitable for effective articulation and animation. Recent deep-learning approaches have also been developed to learn skeletons directly from input shapes without relying on predefined templates. These methods are generally trained on datasets containing thousands of rigged characters, allowing them to generate skeletons that align with articulated components. For instance, Xu et al. \cite{xu2019predicting} introduced a volumetric network designed to generate skeletons for input 3D models. RigNet \cite{xu2020rignet} leverages graph convolutions to learn mesh representations, thereby enhancing the accuracy of skeleton extraction. However, it relies on the strong assumption that the input training and test shapes maintain a consistent upright and front-facing orientation.

In this work, we formulate skeleton generation as an auto-regressive problem to accommodate the varying number of bones in different 3D models. By generating bones auto-regressively, our method dynamically adapts to each model's specific requirements, ensuring flexibility and accuracy in skeleton creation.

\subsection{Skinning weight prediction}
To make 3D models ready for articulation, we also predict skinning weights conditioned on the 3D shape and corresponding skeleton, which define the influence of each joint on each vertex of the mesh.

Several geometric-based techniques have been introduced for skinning \cite{dionne2013geodesic, jacobson2011bounded, dodik2024robust, baran2007automatic}. These methods assign skinning weights based on the distance between joints and vertices. However, this distance-based assumption often fails when the 3D shape has a complex topology. Deep learning-based methods \cite{liu2019neuroskinning, xu2020rignet, liao2022skeleton, mosella2022skinningnet}, such as NeuroSkinning \cite{liu2019neuroskinning}, take a skeleton template as input and predict skinning weights using a learned graph neural network. RigNet \cite{xu2020rignet} utilizes intrinsic shape representations that capture geodesic distances between vertices and bones, often struggles with highly intricate mesh topologies and may require extensive feature engineering to maintain performance across varied object categories. SkinningNet \cite{mosella2022skinningnet} employs a two-stream graph neural network to compute skinning weights directly from input meshes and the corresponding skeletons. However, the performance of these GNN-based methods can degrade when applied to datasets with highly varying orientations, such as \ourdata{}, leading to reduced accuracy and robustness in complex and varied scenarios.

In this work, we predict skinning weights in a functional diffusion process by incorporating volumetric geodesic distance priors between vertices and joints. This approach effectively handles complex mesh topologies and diverse skeletal structures without the constraints of shape orientations.

\begin{figure*}[htbp]
  \centering
  \begin{subfigure}[b]{0.3\textwidth}
    \centering
    \includegraphics[width=\textwidth]{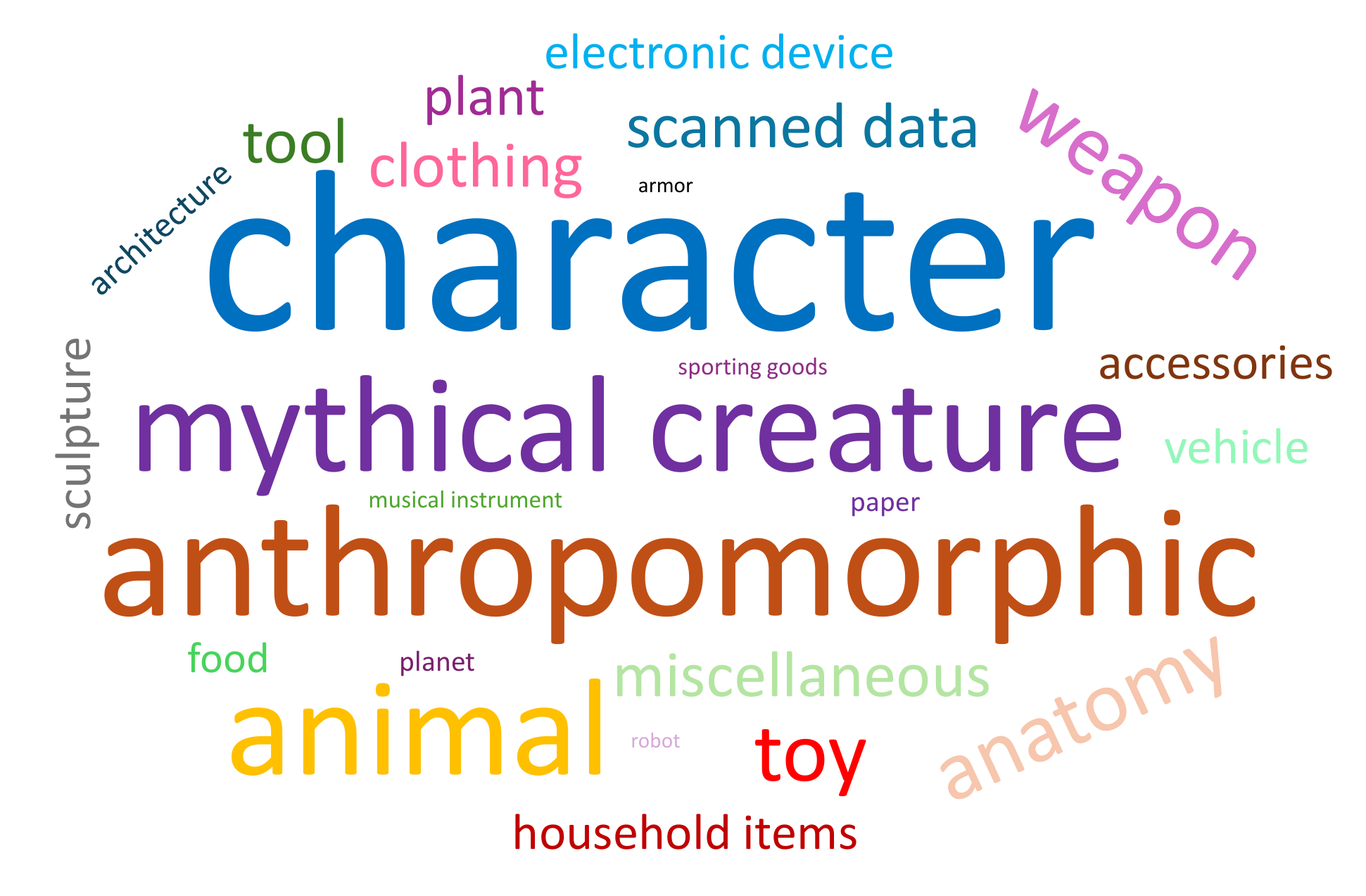}
    \caption{\textbf{Word cloud of \ourdata{} categories.}}
    \label{fig:wordcloud}
  \end{subfigure}
  \hfill
  \begin{subfigure}[b]{0.36\textwidth}
    \centering
    \includegraphics[width=\textwidth]{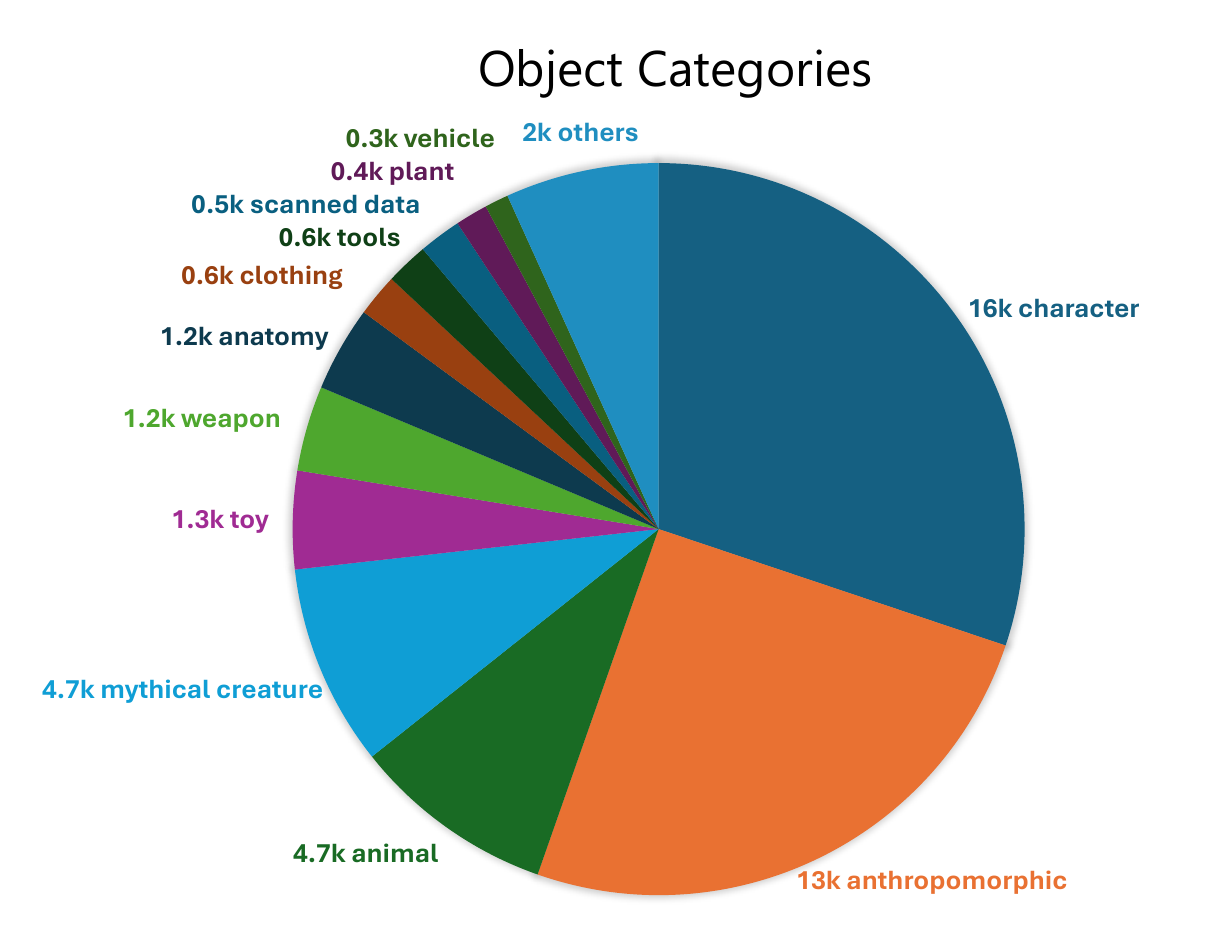}
    \caption{\textbf{Breakdown of \ourdata{} categories.}}
    \label{fig:bing}
  \end{subfigure}
  \hfill
  \begin{subfigure}[b]{0.32\textwidth}
    \centering
    \includegraphics[width=\textwidth]{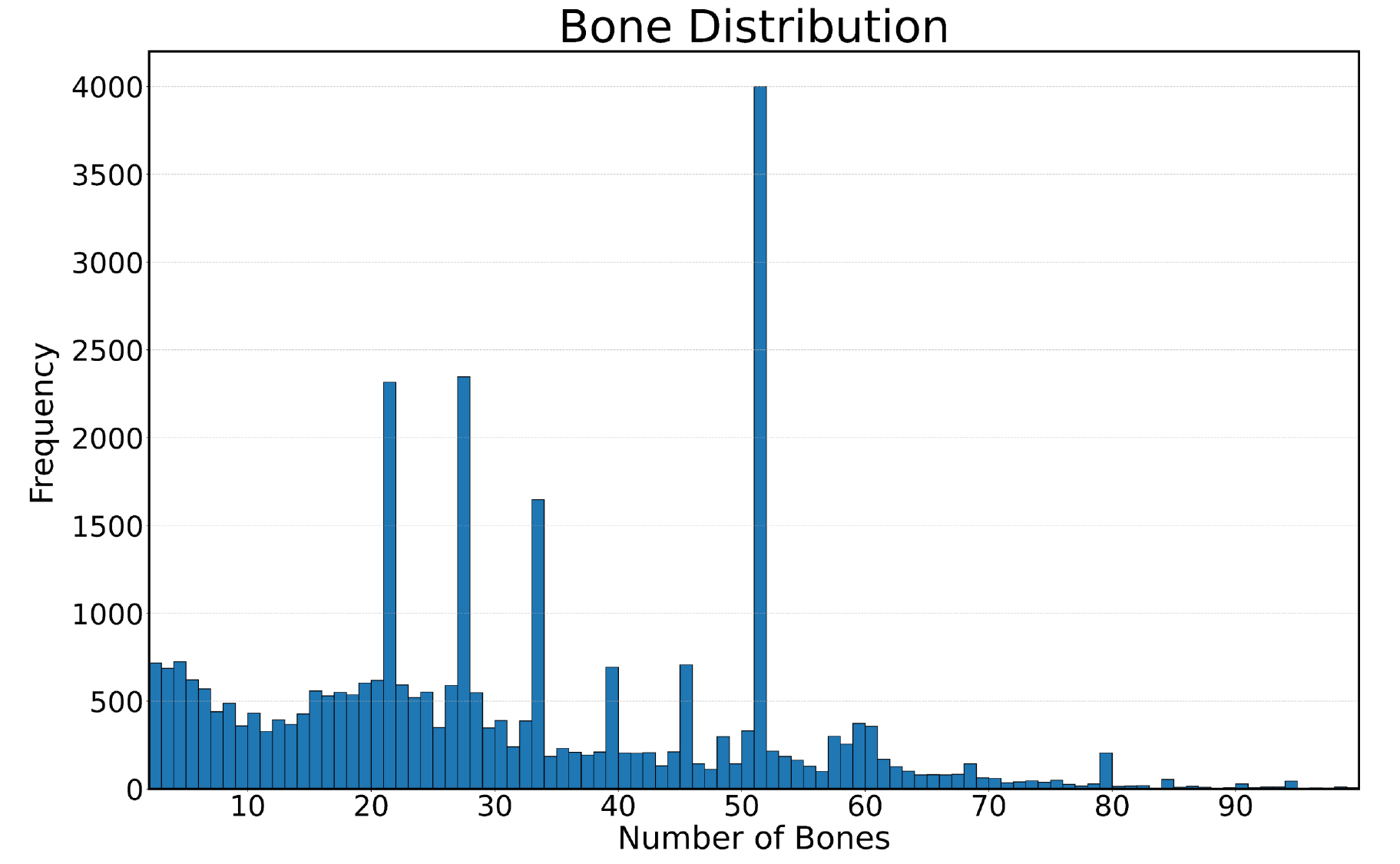}
    \caption{\textbf{Bone number distributions of \ourdata{}.}}
    \label{fig:bone_distributions}
  \end{subfigure}
  
  \caption{\textbf{\ourdata{} statistics.}}
  \label{fig:combined_figure}
  \vspace{-15pt}

\end{figure*}

\subsection{Auto-regressive 3D generation}
Recently, auto-regressive models have been widely used in 3D mesh generation \cite{nash2020polygen, siddiqui2024meshgpt, chen2024meshanything, chen2024meshanythingv2, chen2024meshxl, tang2024edgerunner, weng2024pivotmesh}.
 MeshGPT \cite{siddiqui2024meshgpt} models meshes as sequences of triangles and tokenizes them using a VQ-VAE \cite{van2017neural}. It then employs an auto-regressive transformer to generate the token sequences. This approach enables the creation of meshes with varying face counts. However, most subsequent methods \cite{chen2024meshanything, chen2024meshxl, weng2024pivotmesh} are limited to generating meshes up to 800 faces, due to the computational cost of mesh tokenization. MeshAnythingV2 \cite{chen2024meshanythingv2} introduces Adjacent Mesh Tokenization (AMT), doubling the maximum face count to 1,600. EdgeRunner \cite{tang2024edgerunner} further increases this limit to 4,000 faces by enhancing mesh tokenization techniques. In this work, we explore the potential of auto-regressive models for shape-conditioned skeleton generation. To achieve this, we formulate skeletons as sequences of bones. Unlike mesh generation, which focuses on creating detailed and realistic shapes by utilizing a high number of faces, skeleton generation prioritizes accuracy over complexity. Accurate skeletons are crucial for realistic articulation and animation, and typically consist of fewer than 100 bones, as indicated by the statistics in \ourdata{}.

\section{Articulation-XL}
\label{arti-xl}

\begin{figure}
    \centering
    \includegraphics[scale=0.26]{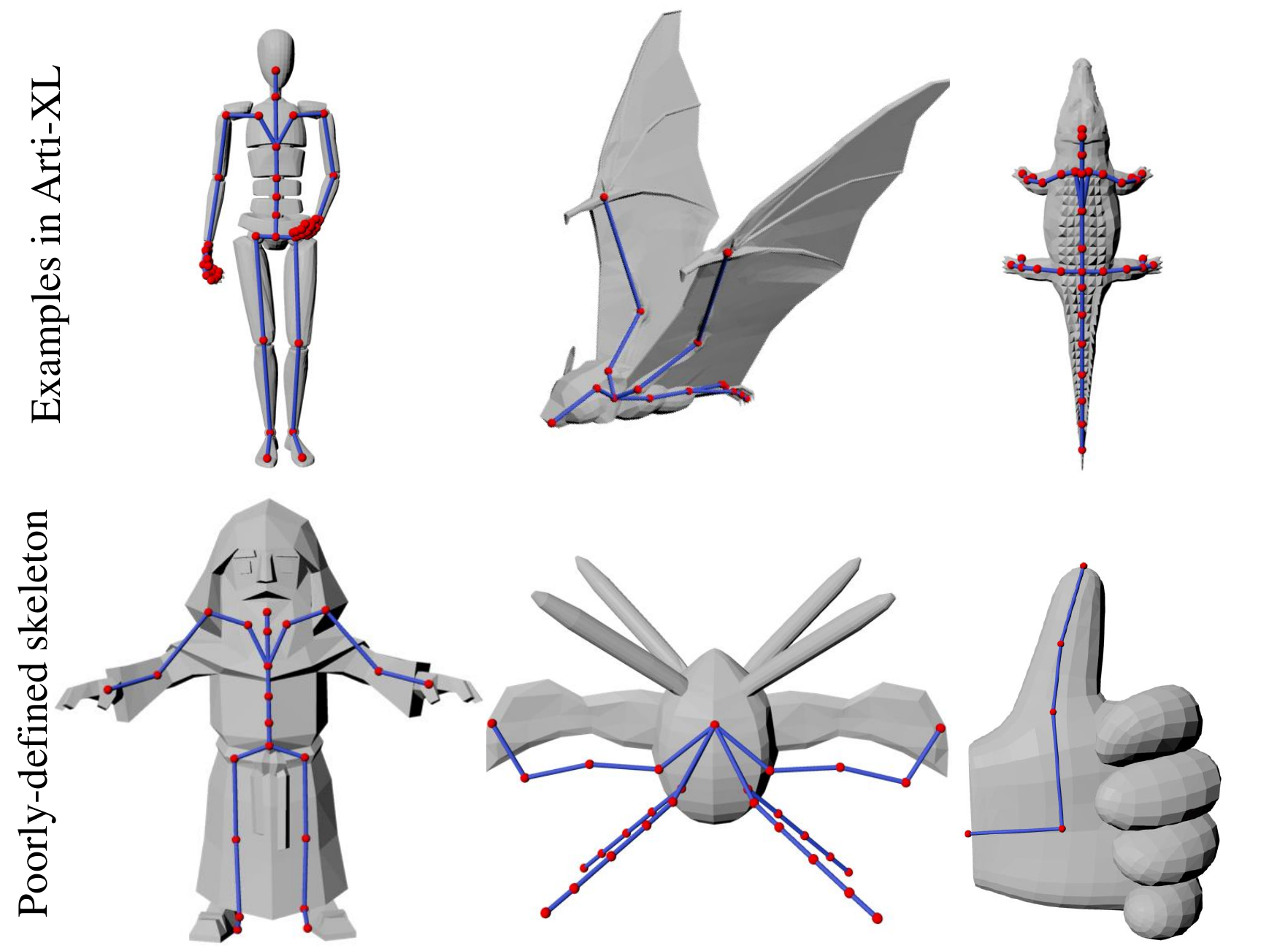}
    \caption{\textbf{Some examples from \ourdata{} alongside examples of poorly defined skeletons that were curated out.}}
    \label{fig:examples}
    \vspace{-16pt}
  \end{figure}

To facilitate large-scale learning of 3D model articulation, we present \ourdata{}, a comprehensive dataset curated from Objaverse-XL \cite{deitke2023objaverse, deitke2024objaverse}. Our dataset construction pipeline consists of three main stages: initial filtering, VLM-based filtering, and category annotation. 
\textbf{We will release our \ourdata{} to facilitate future work.}

\boldstartspace{Initial data collection.} We begin by identifying 3D models from Objaverse-XL that contain both skeleton and skinning weight annotations. To ensure data quality and practical utility, we apply the following filtering criteria: 1) we remove duplicate data based on both skeleton and mesh similarity; 2) we exclude models with only a single joint/bone structure; 3) we filter out data with more than 100 bones, which constitute a negligible portion of the dataset. This initial filtering yields 38.8k candidate models with articulation annotations.

\boldstartspace{VLM-based filtering.} However, we observe that many initial candidates contain poorly defined skeletons that may impair learning (see \Cref{fig:examples}). To ensure dataset quality, we further implement a Vision-Language Model (VLM)-based filtering pipeline: 1) we render each object with its skeleton from four viewpoints; 2) and then utilize GPT-4o \cite{openai_gpt4o} to assess skeleton quality based on specific criteria (detailed in supplementary).
This process results in a final collection of over 33k 3D models with high-quality articulation annotations, forming the curated dataset \ourdata{} \footnote{We have expanded the dataset to over 48K models in Articulation-XL2.0. For further details, please refer to \url{https://huggingface.co/datasets/chaoyue7/Articulation-XL2.0}.}. The dataset exhibits diverse structural complexity: the number of bones per model ranges from 2 to 100, and the number of joints ranges from 3 to 101. The distribution of bone numbers is illustrated in \Cref{fig:bone_distributions}. 

\boldstartspace{Category label annotation.} Additionally, we also leverage a Vision-Language Model (VLM) to automatically assign category labels to each model using specific instructions.
The distribution of these categories is illustrated via a word cloud and a pie chart, as shown in \Cref{fig:wordcloud} and \Cref{fig:bing}, respectively. We observe a rich diversity of object categories, with human-related models forming the largest subset. Detailed statistics and distribution analyses are provided in the supplementary material.

\section{Methods}

\begin{figure*}
  \centering
\includegraphics[width=\textwidth]
  {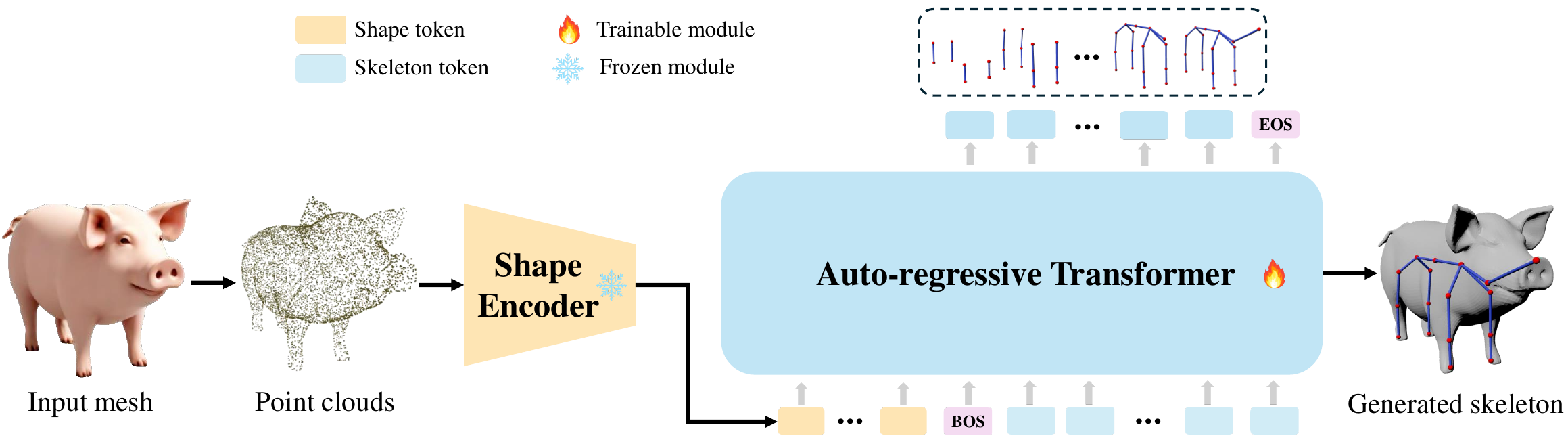}
  \caption{\textbf{Overview of our method for auto-regressive skeleton generation.} Given an input mesh, we begin by sampling point clouds from its surface. These sampled points are then encoded into fixed-length shape tokens, which are appended to the start of skeleton tokens to achieve auto-regressive skeleton generation conditioned on input shapes. The input mesh is generated by Rodin Gen-1 \cite{zhang2024clay}.}
  \label{fig_ar}
\end{figure*}

We propose a two-stage pipeline to make 3D models articulation-ready. Given an input 3D mesh, our method first employs an auto-regressive transformer to generate a structurally coherent skeleton (\Cref{ar}). Subsequently, we predict skinning weights in a functional diffusion process, conditioning on both the input shape and its corresponding skeleton (\Cref{skin}). 

\subsection{Auto-regressive skeleton generation}
\label{ar}
In the initial stage of \ours{}, we generate skeletons for 3D models. Unlike previous approaches that rely on fixed templates, our method can handle the inherent structural diversity of 3D objects through an auto-regressive generation framework, as presented in \Cref{fig_ar}.

\subsubsection{Problem formulation}
Given an input 3D mesh $\mathcal{M}$, our goal is to generate a structurally valid skeleton $\mathcal{S}$ that captures the articulation structure of the object. A skeleton consists of two key components: a set of joints $\mathbf{J} \in \mathbb{R}^{j \times 3}$ defining spatial locations, and bone connections $\mathbf{B} \in \mathbb{N}^{b \times 2}$ specifying the topological structure through joint indices. Formally, we aim to learn the conditional distribution:

\begin{equation} \mathit{p}(\mathcal{S} | \mathcal{M}) = \mathit{p}(\mathbf{J}, \mathbf{B} | \mathcal{M}), \end{equation}
where $\mathcal{M}$ can be sourced from various inputs, including direct 3D models, text-to-3D generation, or image-based reconstruction.

A key challenge in skeleton generation lies in the variable complexity of articulation structures across different objects. Traditional approaches \cite{baran2007automatic, li2021learning} often adopt predefined skeleton templates, which work well for specific categories like human bodies but fail to generalize to objects with diverse structural patterns. 
This limitation becomes particularly apparent when dealing with our large-scale dataset that contains a wide range of object categories.

To address this challenge, we draw inspiration from recent advances in auto-regressive mesh generation \cite{siddiqui2024meshgpt, chen2024meshanythingv2} and reformulate skeleton generation as a sequence modeling task. This novel formulation allows us to:
1) handle varying numbers of bones or joints within skeletons across different 3D models;
2) capture the inherent dependencies between bones;
3) scale effectively to diverse object categories.

\subsubsection{Sequence-based generation framework}
Our framework transforms the skeleton generation task into a sequence modeling problem through four key components: skeleton tokenization, sequence ordering, shape conditioning, and auto-regressive generation. 

\boldstartspace{Skeleton tokenization.}
We represent each skeleton $\mathcal{S}$ as a sequence of bones, where each bone is defined by its two connecting joints ($6$ coordinates in total). To ensure consistent and discrete representation, we employ a carefully designed tokenization process. 
We first scale and translate the input mesh and corresponding skeleton to a unit cube $[-0.5, 0.5]^3$, ensuring their spatial alignment. 
Subsequently, we map the normalized joint coordinates to a discrete $128^3$ space, leading to a sequence length of $6b$ for $b$ bones. 
As such, the discretized coordinates are converted into tokens, which serve as input to the auto-regressive transformer.
Unlike MeshGPT \cite{siddiqui2024meshgpt}, we omit the VQ-VAE compression step based on our dataset analysis. Specifically, in \ourdata{}, most of the models have fewer than 100 bones (i.e., 600 tokens). 
Given these relatively short sequence lengths, using VQ-VAE compression would potentially introduce artifacts without significant benefits in computational efficiency.

\begin{figure}
  \centering
\includegraphics[scale=0.22]
{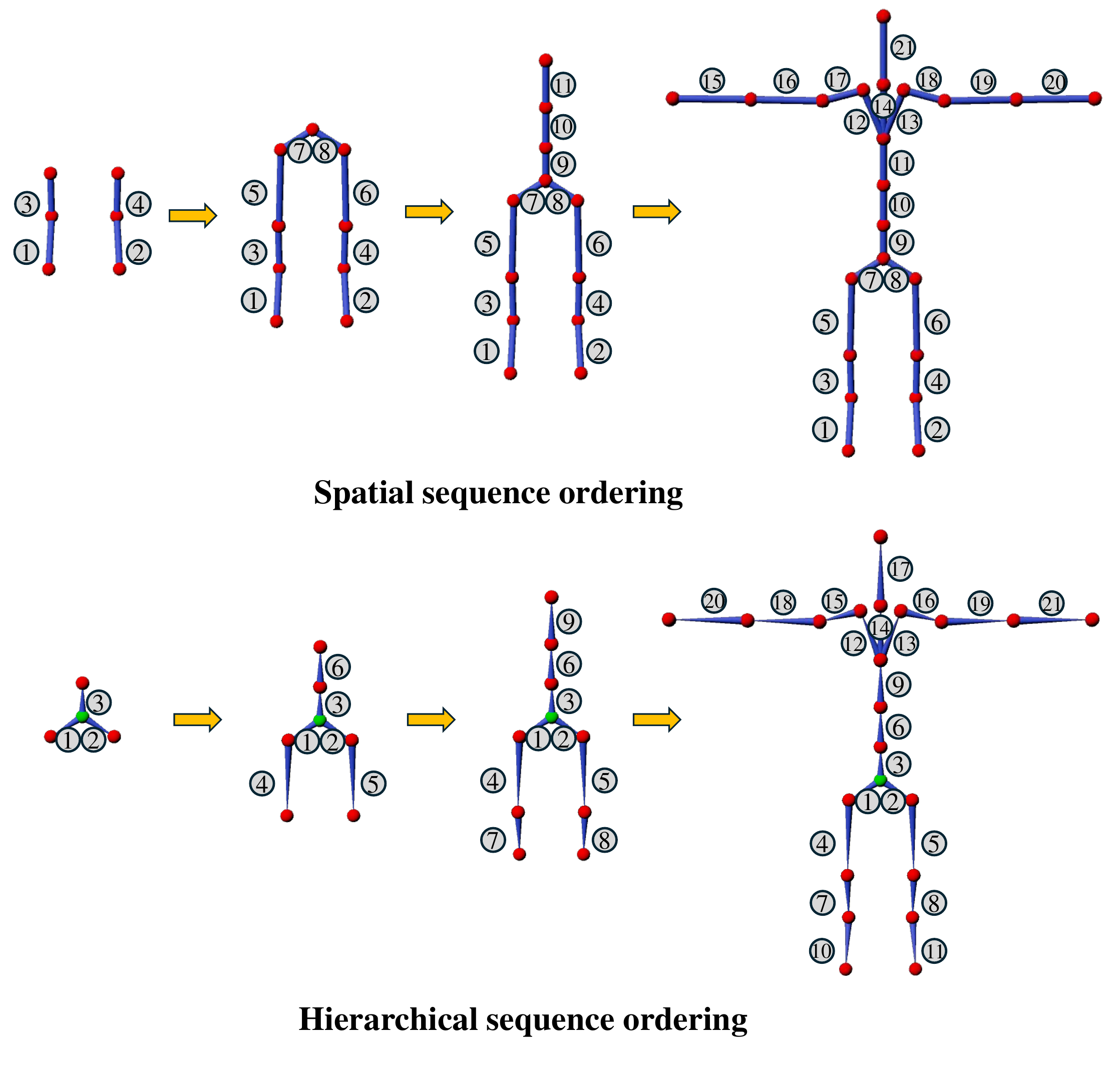}
  \caption{\textbf{Spatial sequence ordering versus hierarchical sequence ordering.} The numbers indicate the bone ordering indices.}
  \label{fig_ar}
\end{figure}

\boldstartspace{Sequence ordering.} 
In this work, we investigate two distinct ordering strategies. Our first approach follows the sequence ordering strategy from recent 3D mesh generation methods \cite{nash2020polygen, siddiqui2024meshgpt}. In this approach, joints are initially sorted in ascending z-y-x order (with z representing the vertical axis), and the corresponding joint indices in the bones are updated accordingly. Bones are then ordered first by their lower joint index and subsequently by the higher one. Additionally, for each bone, the joint indices are cyclically permuted so that the lower index appears first. we refer to this ordering as \textbf{spatial sequence ordering} in this paper. However, this ordering strategy disrupts the parent-child relationships among bones and does not facilitate identifying the root joint. Consequently, additional processing is required to build the skeleton's hierarchy.

To overcome these limitations, we propose an alternative approach termed \textbf{hierarchical sequence ordering}\footnote{Hierarchical ordering is an extension of our under review version.}, which leverages the intrinsic hierarchical structure of the skeleton by processing bones layer by layer. After sorting joints in ascending z-y-x order and updating their indices in bones, we first order the bones directly connected to the root joint. When the root has several child joints, we begin with the bone linked to the child joint having the smallest index and then proceed in ascending order. For subsequent layers, bones are grouped by their immediate parent, and within each group, they are arranged in ascending order based on the child joint index. Additionally, among groups in the same layer, the group corresponding to the smallest parent joint index is processed first, followed by those with larger indices.

\boldstartspace{Shape-conditioned generation.}
Following the conventions in \cite{chen2024meshanythingv2, chen2024meshanything}, we utilize point clouds as the shape condition by sampling 8,192 points from the input mesh $\mathcal{M}$. 
We then process this point cloud through a pre-trained shape encoder \cite{zhao2024michelangelo}, which transforms the raw 3D geometry into a fixed-length feature sequence suitable for transformer processing. 
This encoded sequence is then appended to the start of the transformer's input skeleton sequence for auto-regressive generation. Additionally, for each sequence, we insert a $\textless \mathrm{bos} \textgreater$ token after the shape latent tokens to signify the beginning of the skeleton tokens. Similarly, a $\textless \mathrm{eos} \textgreater$ token is added following the skeleton tokens to denote the end of the skeleton sequence.

\boldstartspace{Auto-regressive learning.}
For skeleton generation, we employ a decoder-only transformer architecture, specifically the OPT-350M model \cite{zhang2022opt}, which has demonstrated strong capabilities in sequence modeling tasks.  During training, we provide the ground truth sequences and utilize cross-entropy loss for next-token prediction to supervise the model: \begin{equation} 
    \mathcal{L}_{pred} = \mathrm{CE}(\mathbf{T}, \mathbf{\hat{T}}) ,
\end{equation} 
where $\mathbf{T}$ represents the one-hot encoded ground truth token sequence, and $\mathbf{\hat{T}}$ denotes the predicted sequence.

At inference time, the generation process begins with only the shape tokens as input, and the model sequentially generates each skeleton token, ending when the $\textless \mathrm{eos} \textgreater$ token is produced. The resulting token sequence is then detokenized to recover the final skeleton coordinates and connectivity structure.

\subsection{Skinning weight prediction}
\label{skin}

The second stage focuses on predicting skinning weights, which controls how the mesh deforms with skeleton movements. In this work, we represent skinning weights as an $n$-dimensional function defined on mesh surfaces, which are continuous, high-dimensional, and exhibit significant variation across different skeletal structures. To address these complexities, we employ a functional diffusion framework for accurate skinning weight prediction.

\subsubsection{Preliminary: Functional diffusion}

Functional diffusion \cite{zhang2024functional} extends classical diffusion models to operate directly on functions, making it particularly suitable for our task. Consider a function $f_0$ mapping from domain $\mathcal{X}$ to range $\mathcal{Y}$:
\begin{equation}
f_0 : \mathcal{X} \rightarrow \mathcal{Y}.
\end{equation}

The diffusion process gradually adds functional noise $g$ (mapping the same domain to range) to the original function:
\begin{equation}
f_t(x) = \alpha_t \cdot f_0(x) + \sigma_t \cdot g(x), \quad t \in [0, 1]
\end{equation}
where $\alpha_t$ and $\sigma_t$ control the noise schedule. The goal is to train a denoiser $D$ that recovers the original function:
\begin{equation}
D_\theta[f_t, t](x) \approx f_0(x).
\end{equation}

This formulation naturally aligns with our task requirements. By treating skinning weights as continuous functions over the mesh surface, we can capture smoothly transitioning weights between vertices. Additionally, the framework's flexibility allows it to adapt to diverse mesh topologies and skeletal structures.

\subsubsection{Skinning weight prediction}
Building upon the functional diffusion framework, we formulate skinning weight prediction as learning a mapping \( f: \mathbb{R}^3 \rightarrow \mathbb{R}^n \) from 3D points to their corresponding weights. Specifically, the input to our model consists of 3D points \( \mathcal{P} \in \mathbb{R}^{v \times 3} \) sampled from the surface of the mesh. The output is an \( n \)-dimensional skinning weight matrix \( \mathcal{W} \in \mathbb{R}^{v \times n} \). Here, the ground truth skinning weights of sampled points for training are copied from their nearest vertices and will also be copied back when inference. $n$ denotes the maximum number of joints in the dataset.

To enhance prediction accuracy, we introduce two key components. 
First, we condition the generation on both joint coordinates and global shape features extracted by a pre-trained encoder \cite{zhao2024michelangelo}. 
Second, we leverage volumetric geodesic priors calculated from \cite{dionne2013geodesic}. Specifically, we compute the volumetric geodesic priors from each mesh vertex to each joint. We then assign these priors to sampled points based on their nearest vertices and normalize them to match the range of skinning weights, forming a volumetric geodesic matrix $\mathcal{G} \in \mathbb{R}^{v \times n}$. Our model learns to predict the residual between the actual skinning weights and this geometric prior, i.e., $f: \mathcal{P} \rightarrow (\mathcal{W}-\mathcal{G})$, enabling more stable predictions.

Following \cite{zhang2024functional}, we optimize our model using $x_0$-prediction with the objective:
\begin{equation}
   \mathcal{L}_{denoise} = \left\| D_{\theta} \left( \left\{ x, f_t(x) \right\}, t \right) - f_0(x) \right\|^2_2, \quad x \in \mathcal{P}.
\end{equation}
We employ the Denoising Diffusion Probabilistic Model (DDPM) \cite{ho2020denoising} as our scheduler. In practice, we normalize the skinning weights and volumetric geodesic priors to the range \([-1, 1]\) before adding noise. We will conduct ablation studies on this design in \Cref{ablate_skin}.

\section{Experiments}
\subsection{Implementation details}
\boldstartspace{Datasets.} We evaluate our method on two datasets: our proposed \ourdata{} and ModelsResource \cite{xu2020rignet, ModelsResource2019}.  \ourdata{} contains 33k samples, with 31.4k for training and 1.6k for testing. ModelsResource is a smaller dataset, containing 2,163 training and 270 testing samples. The number of joints for each object varies from 3 to 48, with an average of 25.0 joints. While the data in \res{} maintains a consistent upright and front-facing orientation, the 3D models in \ourdata{} exhibit varying orientations. We have verified that there are no duplications between \ourdata{} and ModelsResource.

\boldstartspace{Training details.} Our training process consists of two stages. For skeleton generation, we train the auto-regressive transformer on 8 NVIDIA A100 GPUs for approximately two days. For skinning weight prediction, models are trained on the same hardware configuration for about one day. To enhance model robustness, we apply data augmentation including scaling, shifting, and rotation transformations. For more details, please refer to the appendix.

\subsection{Skeleton generation results}
\boldstartspace{Metrics.}
We adopt three standard metrics following \cite{xu2020rignet} to evaluate skeleton quality: CD-J2J, CD-J2B, and CD-B2B. 
These Chamfer Distance-based metrics measure the spatial alignment between generated and ground truth skeletons by computing distances between joints-to-joints, joints-to-bones, and bones-to-bones respectively. Lower values indicate better skeleton quality.

\boldstartspace{Baselines.} 
We compare our method against two representative approaches: Pinocchio \cite{baran2007automatic}, a traditional template-fitting method, and RigNet \cite{xu2020rignet}, a learning-based method using graph convolutions. All methods are evaluated on the \ourdata{} and ModelsResource datasets.

\begin{figure*}
  \centering
  \includegraphics[width=\textwidth]{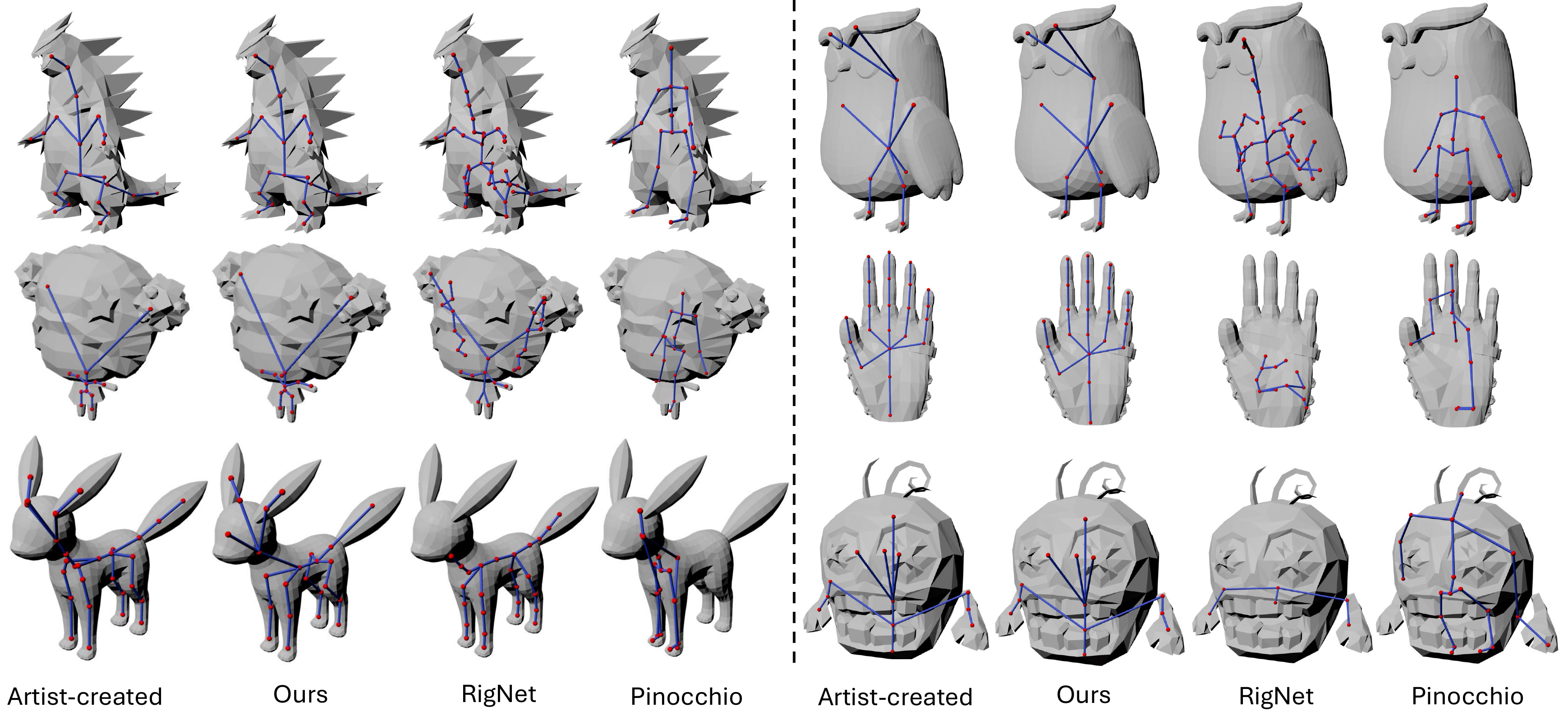}
  \vspace{-10pt}
  \caption{\textbf{Comparison of skeleton creation results on \res{} (left) and \ourdata{} (right).} Our generated skeletons more closely resemble the artist-created references, while RigNet and Pinocchio struggle to handle various object categories. }
  \label{compare_skel}
  \vspace{-15pt}
\end{figure*}

\boldstartspace{Comparison results.}
Qualitative comparisons are presented in \Cref{compare_skel}, where we compare different methods across various object categories.  
Pinocchio struggles with objects that differ from its predefined templates, especially obvious in non-humanoid objects (as shown in the 2nd row and the 3rd row on the right).
RigNet demonstrates improved performance when tested on \res{}, where the data maintains a consistent upright and front-facing orientation. However, it still struggles with complex topologies (as illustrated in the 1st and 2nd rows on the left). Furthermore, RigNet performs worse on \ourdata{}, where the data exhibit varying orientations. 
In contrast, our method generates high-quality skeletons that closely match artist-created references across diverse object categories.

The quantitative results are shown in \Cref{comparison_skel}. Our method consistently outperforms baselines across all metrics on both datasets. Additionally, we compare our method using both spatial and hierarchical ordering strategies. The spatial ordering consistently achieves better performance, likely because the hierarchical ordering requires the model to allocate part of its capacity to learning the skeleton’s hierarchy and identifying the root joint. Results obtained using spatial ordering are well-suited for applications such as skeleton-driven pose transfer \cite{zhang2024magicpose4d}, whereas those derived from hierarchical ordering are more readily integrated with 3D models for animation.

\begin{table}
  \caption{\textbf{Quantitative comparison on skeleton generation.} We compare different methods using CD-J2J, CD-J2B, and CD-B2B as evaluation metrics on both \ourdata{} (Arti-XL) and ModelsResource (Modelres.). Lower values indicate better performance. The metrics are in units of $10^{-2}$. Here, * denotes models trained on \ourdata{} and tested on ModelsResource.}
  \vspace{-8pt}
  \label{comparison_skel}
  \centering
  \begin{tabular}{ccccc}
    \toprule
      & Dataset &  CD-J2J  & CD-J2B & CD-B2B \\
    \midrule
    RigNet*  & \multirow{7}{*}{\textit{ModelsRes.}} & 7.132 & 5.486 & 4.640 \\
    Pinocchio  &  & 6.852 & 4.824 & 4.089 \\
    Ours-hier*  &  & 4.451 & 3.454 & 2.998 \\
    RigNet  &  & 4.143 & 2.961  & 2.675 \\
    Ours-spatial*  &  & 4.103 & 3.101 & 2.672 \\
    Ours-hier     &   & 3.654 & 2.775 & 2.412     \\
    Ours-spatial     &   & \textbf{3.343} & \textbf{2.455} & \textbf{2.140}      \\
    
    \midrule
    Pinocchio  &\multirow{4}{*}{\textit{Arti-XL}} & 8.360 & 6.677  & 5.689 \\
    RigNet  & & 7.478 & 5.892 & 4.932   \\
    Ours-hier   &   & 3.025 & 2.408 & 2.083      \\
    Ours-spatial   &   & \textbf{2.586} & \textbf{1.959} & \textbf{1.661}      \\
    
    \bottomrule
  \end{tabular}
  \vspace{-15pt}
\end{table}

\boldstartspace{Generalization analysis.} 
To evaluate the generalization capability, we 
perform cross-dataset evaluation by training RigNet and our MagicArticulate on Articulation-XL and testing on ModelsResource.
As shown in \Cref{comparison_skel} (marked with \textbf{*}), our method maintains competitive performance compared to RigNet trained directly on ModelsResource, while RigNet's performance degrades significantly when tested on unseen data distributions, performing even worse than the template-based method Pinocchio.

\begin{figure}
  \centering
  \includegraphics[scale=0.28]{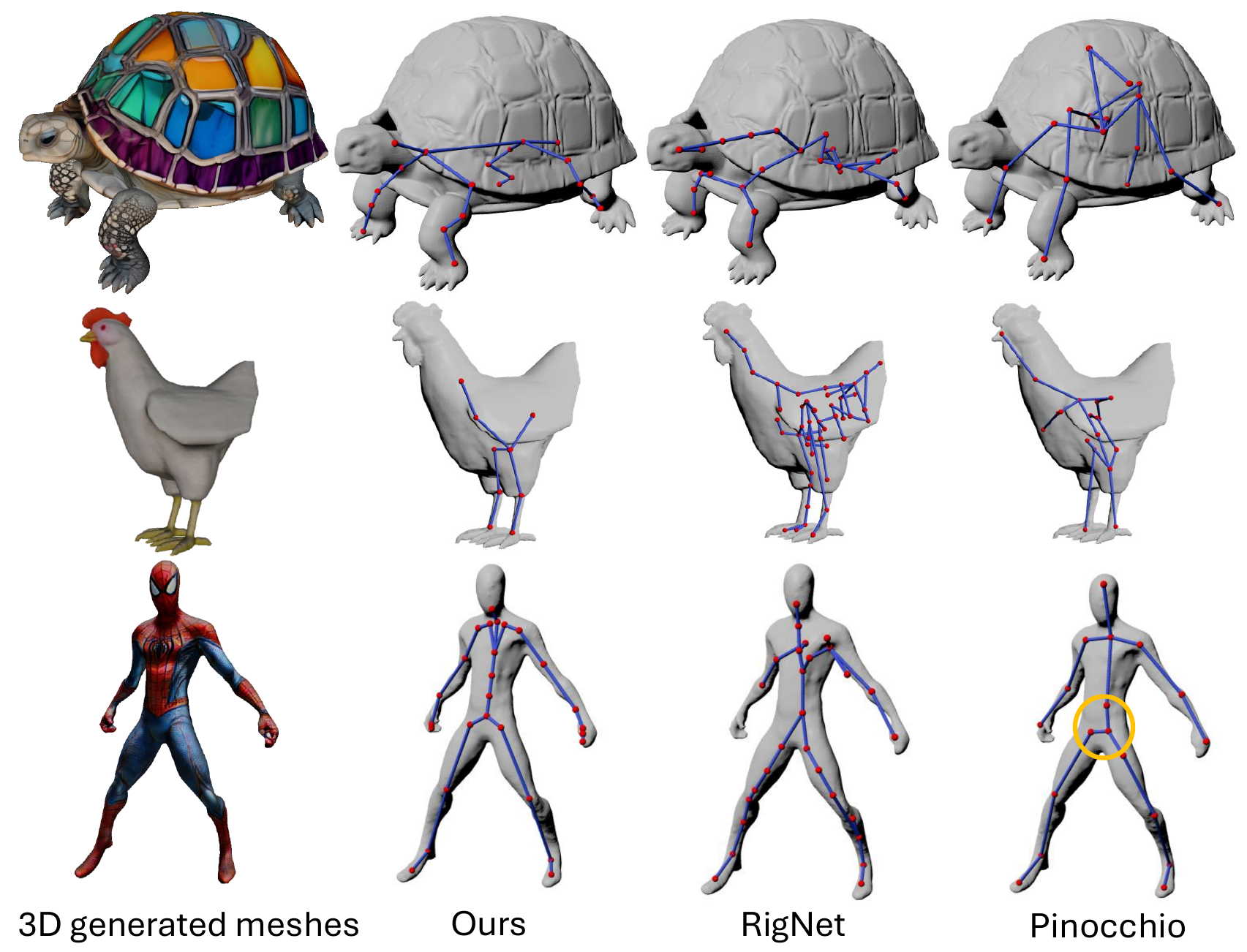}
  \caption{\textbf{Skeleton creation results on 3D generated meshes.} Our method has a better generalization performance than both RigNet \cite{xu2020rignet} and Pinocchio \cite{baran2007automatic} across difference object categories. The 3D models are generated by Tripo 2.0 \cite{tripo3d}.}
  \label{generalization}
\end{figure}

To further assess real-world applicability, we evaluate all methods on AI-generated 3D meshes from Tripo 2.0 \cite{tripo3d} (\Cref{generalization}). 
Our method successfully generates plausible skeletons for diverse object categories, while RigNet fails to produce valid results despite being trained on our large-scale dataset. 
Notably, even Pinocchio's template-based approach struggles to generate accurate skeletons for basic categories like humans and quadrupeds, highlighting the advantage of our method in handling novel object structures.

\begin{figure*}
  \centering
  \includegraphics[width=\textwidth]{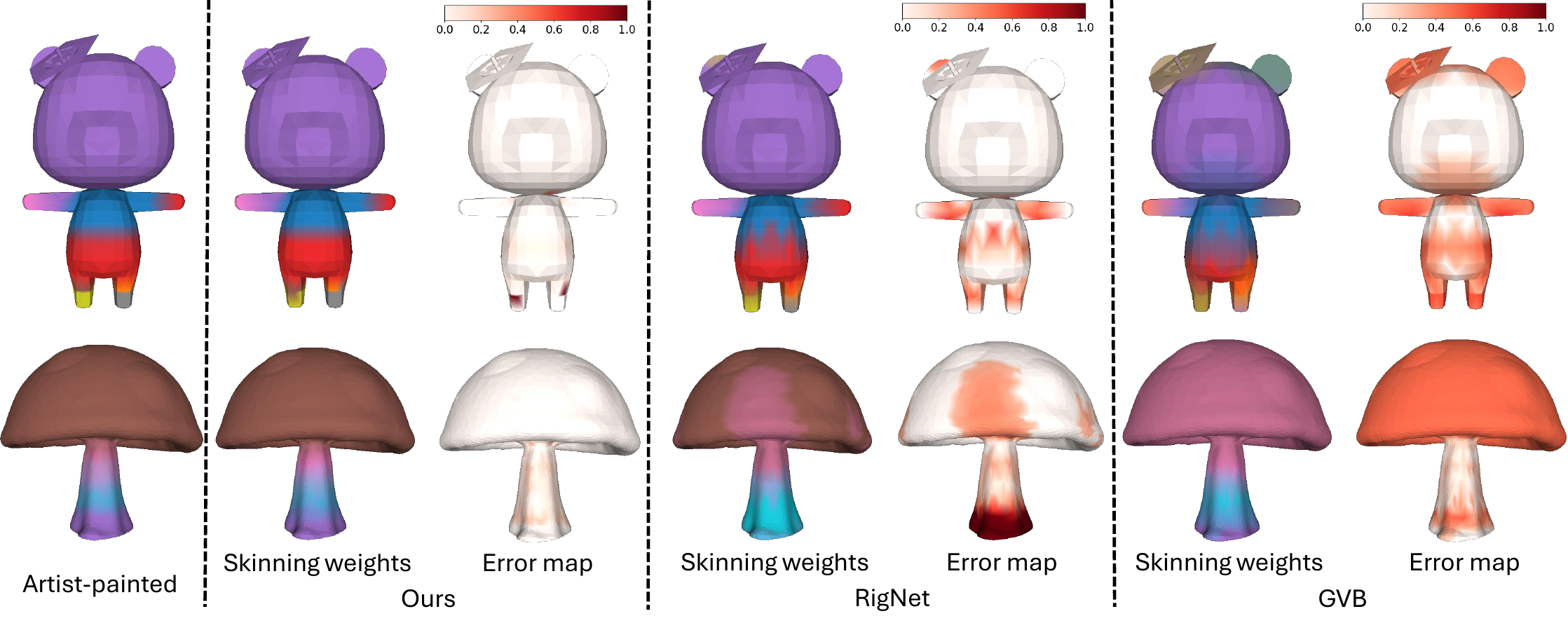}
  \caption{\textbf{Comparisons with previous methods for skinning weight prediction on \res{} (top) and \ourdata{} (bottom).} We visualize skinning weights and L1 error maps. For more results, please refer to the supplementary materials.}
  \label{compare_skin}
\end{figure*}

\subsection{Skinning weight prediction results}
\boldstartspace{Metrics.} 
We evaluate skinning weight quality using three metrics: precision, recall, and L1-norm error.
Precision and recall measure the accuracy of identifying significant joint influences (defined as weights larger than $1e-4$ following \cite{xu2020rignet}, while the L1-norm error computes the average difference between predicted and ground truth skinning weights across all vertices. We will also report the deformation error in appendix.

\boldstartspace{Baselines.}
We compare our method against Geodesic Voxel Binding (GVB) \cite{dionne2013geodesic}, a geometric-based method available in Autodesk Maya \cite{AutodeskMaya2024} and RigNet \cite{xu2020rignet}. 
When trained on \ourdata{}, we filter out a subset containing 28k training and 1.2k testing samples, excluding data with more than 55 joints (which constitute a small fraction of both real-world cases and \ourdata{}).

\boldstartspace{Comparison results.}
Qualitative comparisons in \Cref{compare_skin} visualize the predicted skinning weights and their L1 error maps against artist-created references. 
Our method predicts more accurate skinning weights with significantly lower errors across diverse object categories.
In contrast, both GVB and RigNet show larger deviations, particularly in regions around joint boundaries. 

The quantitative results are shown in \Cref{comparison_skin}, which support qualitative observations,
demonstrating that our method consistently outperforms baselines across most metrics on both datasets.

\begin{table}
  \caption{\textbf{Quantitative comparison on skinning weight prediction.} We compare our method with GVB and RigNet. For Precision and Recall, larger values indicate better performance. For average L1-norm error, smaller values are preferred. }
  \vspace{-10pt}
  \label{comparison_skin}
  \centering
  \begin{tabular}{ccccc}
    \toprule
      & Dataset &  Precision  & Recall & avg L1 \\
    \midrule
    GVB &  \multirow{3}{*}{\textit{ModelsResource}}  & 69.3\%  & 79.2\%  & 0.687    \\
    RigNet  &  & 77.1\% & \textbf{83.5\%}  & 0.464  \\
    Ours     &   & \textbf{82.1{\%}} & 81.6{\%} & \textbf{0.398}      \\
    \midrule
    GVB  &\multirow{3}{*}{\textit{\ourdata{}}} & 75.7\% & 68.3\% & 0.724  \\
    RigNet & & 72.4\% & 71.1\%& 0.698    \\
    Ours   &   & \textbf{80.7{\%}} & \textbf{77.2{\%}} & \textbf{0.337}      \\
    \bottomrule
  \end{tabular}
  \vspace{-12pt}
\end{table}

\vspace{-5pt}
\subsection{Ablation studies}

\subsubsection{Ablation studies on skeleton generation}
We conduct ablation studies to assess the impact of VLM-based data filtering and the number of sampled mesh points on skeleton generation. 
The results, presented in \Cref{ablation_skel}, show notable performance degradation without data filtering, highlighting the importance of high-quality training data. We also vary the number of sampled points as input to the pre-trained shape encoder \cite{zhao2024michelangelo}. As shown in \Cref{ablation_skel}, sampling 8,192 points yields superior performance.

\begin{table}
  \caption{\textbf{Ablation studies for skeleton generation.}}
  \label{ablation_skel}
  \centering
  \begin{tabular}{cccc}
    \toprule
      &  CD-J2J  & CD-J2B & CD-B2B \\
    \midrule
    w/o data filtering   & 2.982 & 2.327  & 2.015 \\
    \midrule
    4,096 points & 2.635 & 2.024  & 1.727 \\
    12,288 points & 2.685 & 2.048  & 1.760 \\
    Ours (8,192)      & \textbf{2.586} & \textbf{1.959} & \textbf{1.661}   
    \\
    \bottomrule
  \end{tabular}
\end{table}

\subsubsection{Ablation studies on skinning weight prediction}
\label{ablate_skin} 
We conduct ablation studies on three critical components of our skinning weight prediction framework. The quantitative results on \res{} are shown in \Cref{ab_skin}. First, removing the volumetric geodesic distance initialization reduces precision by 0.6\% and recall by 3.9\%, demonstrating its crucial role in guiding accurate weight distribution. Second, eliminating our normalization strategy, which scales both skinning weights and geodesic distances to \([-1, 1]\) before noise addition, leads to an 8.7\% increase in L1 error. Finally, excluding global shape features from the pre-trained encoder \cite{zhao2024michelangelo} results in less accurate predictions. All these results validate our design choices and show that each component contributes notably to the final performance.

\begin{table}[]
  \caption{\textbf{Ablation studies on skinning weight prediction.}}
  \label{ab_skin}
  \centering
  \begin{tabular}{cccc}
    \toprule
      &  Precision  &Recall & avg L1  \\
    \midrule
    w/o geodesic dist.  & 81.5\% & 77.7\% & 0.444 \\
    w/o weights norm   & 82.0\% & 77.9\%  & 0.436 \\
     w/o shape features   & 81.4\% &
81.3\% &
0.412 \\
    Ours      & \textbf{82.1{\%}} & \textbf{81.6{\%}} & \textbf{0.398}  
    \\
    \bottomrule
  \end{tabular}
  \vspace{-15pt}
\end{table}

\section{Conclusion}

In this work, we present \ours{} to convert static 3D models into articulation-ready assets that support realistic animation. We first introduce a large-scale dataset \ourdata{} with high-quality articulation annotations, which is carefully curated from Objaverse-XL. Built upon this dataset, we develop a novel two-stage pipeline that first generates skeletons through auto-regressive sequence modeling, naturally handling varying numbers of bones or joints within skeletons across different 3D models. Then we predict skinning weights in a functional diffusion process that incorporates volumetric geodesic distance priors between vertices and joints. 
Extensive experiments demonstrate our method's superior performance and generalization ability across diverse object categories.

\section*{Acknowledgements} 
This research is supported by the MoE AcRF Tier 2 grant (MOE-T2EP20223-0001).
\clearpage
\maketitlesupplementary

\renewcommand{\thetable}{S\arabic{table}}
\renewcommand{\thefigure}{S\arabic{figure}}

\section*{Overview}
In this supplementary material, we provide additional details and experimental results for the main paper, including:

\begin{itemize}
    \item Further details of \ours{} (\Cref{method_detail}) and \ourdata{} (\Cref{detail_data});
    \item Additional experimental results on skeleton generation and skinning weight prediction (\Cref{additioanl_results});
    \item A discussion of the limitations of our work and future works (\Cref{limit}).
\end{itemize}

\section{More details of \ours{}}
\label{method_detail}

\subsection{Implementation details}
\label{implement_detail}
\boldstartspace{Skeleton generation.}
Our skeleton generation pipeline utilizes a pre-trained shape encoder \cite{zhao2024michelangelo} to process input meshes. For each mesh, we sample 8,192 points which are encoded into 257 shape tokens following MeshAnything \cite{chen2024meshanything}. To ensure consistent point cloud sampling across different data sources, we first extract the signed distance function from input mesh using \cite{wang2022dual}, followed by generating a coarse mesh via Marching Cubes \cite{lorensen1998marching}. We then sample point clouds and their corresponding normals from this coarse mesh.

For training on \ourdata{}, we use 8 NVIDIA A100 GPUs for approximately two days with a batch size of 64 per GPU, resulting in an effective batch size of 512. 
When training on \res{}, we utilize 4 NVIDIA A100 GPUs for about 9 hours with a batch size of 32 per GPU, which yields an effective batch size of 128.
During inference, the model generates skeleton tokens auto-regressively from shape tokens until reaching the $\textless \mathrm{eos} \textgreater$ token, followed by detokenization to recover the final skeleton coordinates in $[-0.5, 0.5]$ range.

\begin{figure*}
    \centering
    \includegraphics[scale=0.5]
{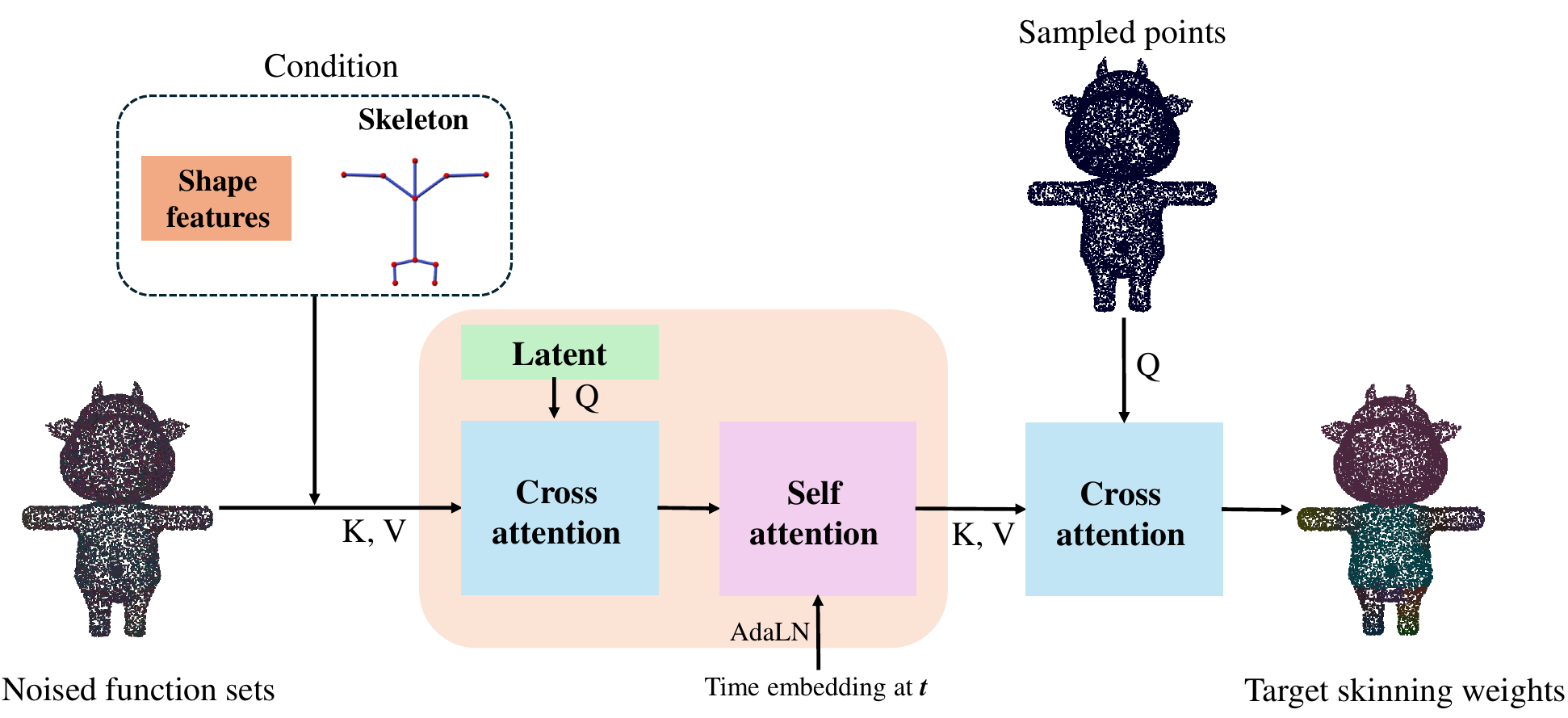}
\caption{\textbf{Overview of the function diffusion architecture for skinning weight prediction.} Given a set of noised skinning weight functions $\{(x, f_{t}(x)) \mid x \in \mathcal{P}\}$, conditioned on skeleton and shape features from \cite{zhao2024michelangelo}, we denoise the skinning weight functions to approximate the target weights.}
    \label{supp_func_network}
  \end{figure*}

\begin{figure}
    \centering
    \includegraphics[scale=0.3]
{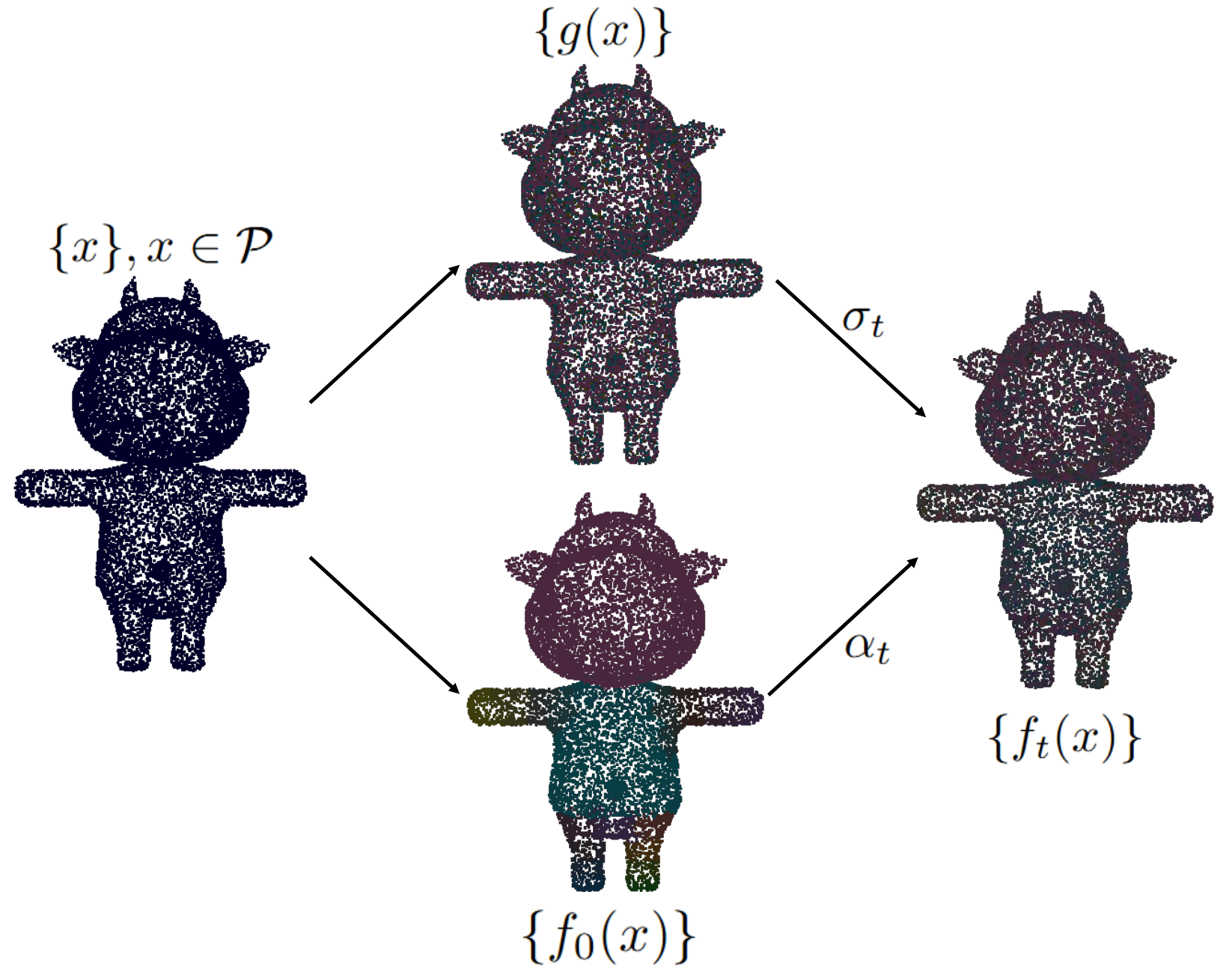}
\caption{\textbf{Process of adding noise to the skinning weight function.} Given $x \in \mathcal{P}$ and the original skinning weight function $f_{0}(x)$, we add the noise function $g(x)$ to obtain the noised function $f_{t}(x)$.}

    \label{supp_func_process}
  \end{figure}

\boldstartspace{Skinning weight prediction.} 
Our functional diffusion model employs the Denoising Diffusion Probabilistic Model (DDPM) with 1,000 timesteps and a linear beta schedule. 
During training, we condition the model on ground truth skeletons and supervise it with corresponding ground truth skinning weights. We add noise to the skinning weight function (the process is illustrated in \Cref{supp_func_process}) and then feed the noised skinning weights into our denoising network (\Cref{supp_func_network}). Following \cite{zhang2024functional}, our network architecture processes the noised set $\{(x, f_{t}(x)) \mid x \in \mathcal{P}\}$ by splitting it into smaller subsets and handling them through multiple cross-attention stages.
The time embedding at timestep $t$ is incorporated into each self-attention layer via adaptive layer normalization. For visual clarity, \Cref{supp_func_network} shows only one processing stage.

We train the model on \ourdata{}  using 8 NVIDIA A100 GPUs for approximately one day, with a batch size of 16 per GPU (effective batch size 128). Training on \res{} uses the same configuration for about 4 hours. During inference, we perform 25 denoising steps to generate predictions $\mathcal{W} \in \mathbb{R}^{v \times n}$ in the range $[-1, 1]$. These results are then normalized to $[0, 1]$, ensuring that each row of the skinning weight matrix sums to 1. To handle varying joint counts across different models, we employ a valid joint mask during both training and testing, with a maximum joint count of 55 as discussed in the main paper (Sections 4.2 and 5.3).

\subsection{Experimental details}

For baseline comparisons, we use the implementations of RigNet \cite{xu2020rignet} and Pinocchio \cite{baran2007automatic} from the GitHub repositories\footnote{\url{https://github.com/zhan-xu/RigNet}, \url{https://github.com/haoz19/Automatic-Rigging}}. 
The Geodesic Voxel Binding (GVB) \cite{dionne2013geodesic} comparison is conducted using the implementation in Autodesk Maya \cite{AutodeskMaya2024}. When training RigNet on our \ourdata{}, we strictly follow the authors' data processing pipeline and six-stage training strategy as specified in their official implementation.

\subsection{Animation}
Many recent works have explored 3D animation, including skeleton-free pose transfer \cite{song20213d, song2023unsupervised, liao2022skeleton}, skeleton-driven pose transfer \cite{zhang2024magicpose4d}, and physics-driven animation \cite{fu2024sync4d}. In this paper, we propose a method that enables automatic articulation generation for any input 3D model, whether artist-created or AI-generated. The pipeline first generates a skeleton for the input model, then predicts skinning weights conditioned on both the model geometry and the generated skeleton. The resulting articulated model can be exported in standard formats (e.g., FBX, GLB), making it directly compatible with popular animation software such as Blender \cite{Blender} and Autodesk Maya \cite{AutodeskMaya2024}.

\section{Additional experimental results}
\label{additioanl_results}
\subsection{More results of skeleton generation}

\begin{figure*}
    \centering
    \includegraphics[scale=0.46]
{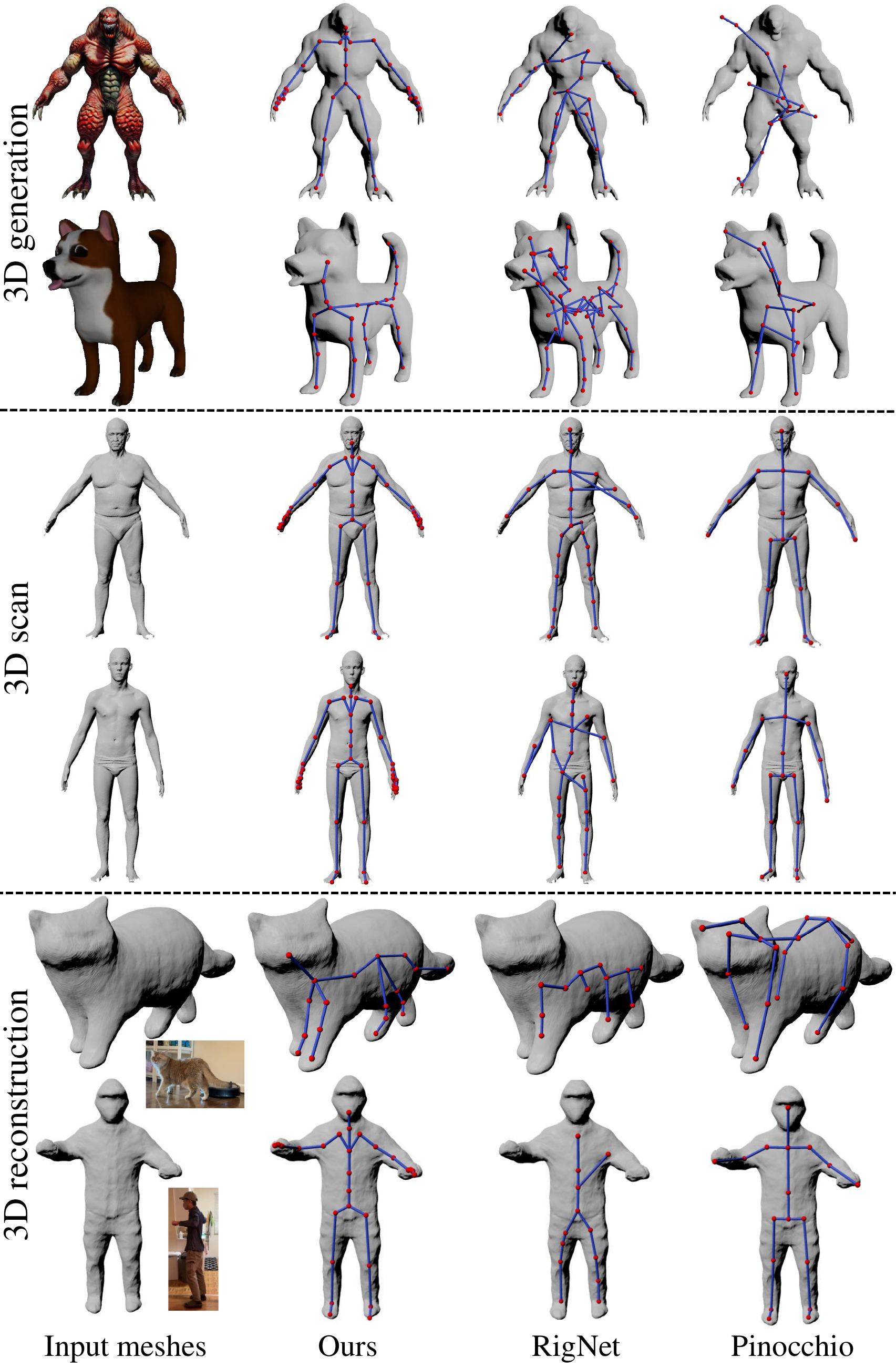}
\caption{\textbf{Comparison of skeleton generation methods on out-of-domain data.} The input meshes are from 3D generation, 3D scan, and 3D reconstruction.}
    \label{supp_skel_ood}
  \end{figure*}
  
\begin{figure*}
    \centering
\includegraphics[width=\textwidth]
    {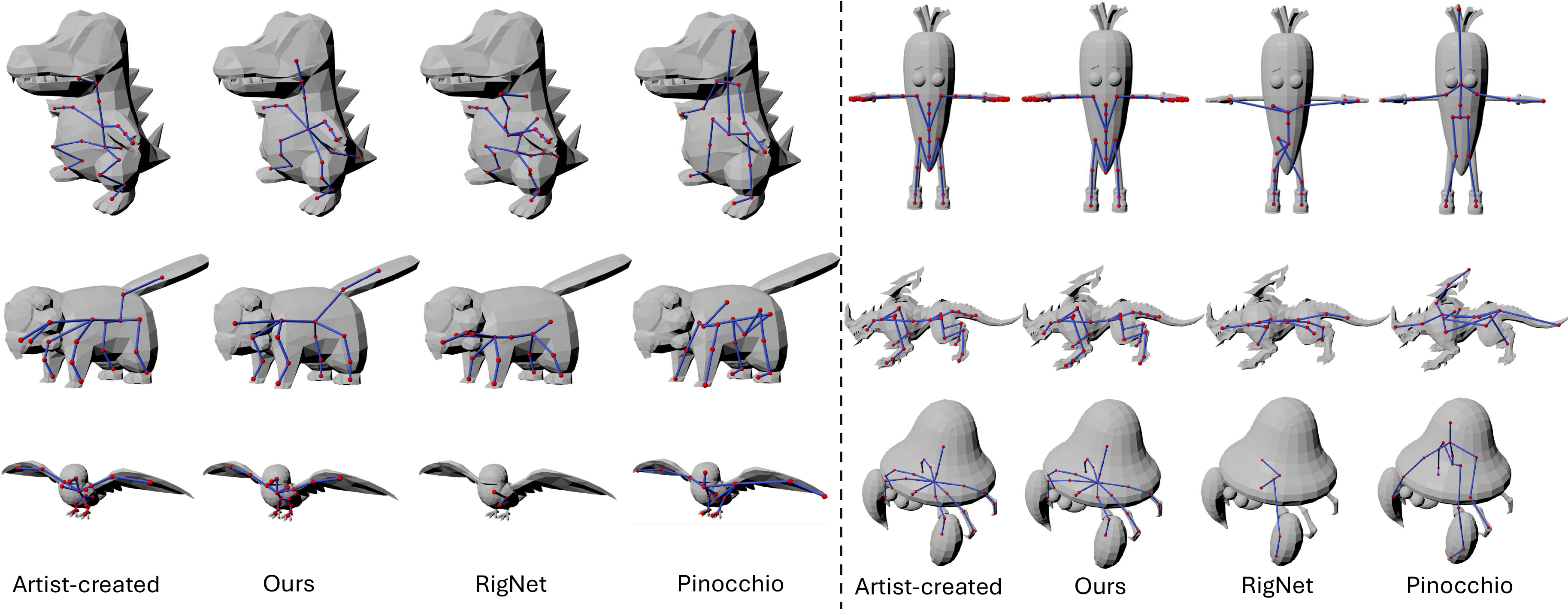}

    \caption{\textbf{Comparison of skeleton generation methods on \res{} (left) and \ourdata{} (right).} Our results more closely resemble the artist-created references, while RigNet and Pinocchio struggle to handle various object categories.}
    \label{supp_skel}
  \end{figure*}

\begin{table*}
  \caption{\textbf{Quantitative comparison on skinning weight prediction.} We compare our method with GVB and RigNet. For Precision and Recall, larger values indicate better performance. For average L1-norm error and average distance error, smaller values are preferred. }
  \label{comparison_skin_supp}
  \centering
  \begin{tabular}{cccccc}
    \toprule
      & Dataset &  Precision  & Recall & avg L1 & avg Dist. \\
    \midrule
    GVB &  \multirow{3}{*}{\textit{ModelsResource}}  & 69.3\%  & 79.2\%  & 0.687  & 0.0067  \\
    RigNet  &  & 77.1\% & \textbf{83.5\%}  & 0.464 & 0.0054 \\
    Ours     &   & \textbf{82.1{\%}} & 81.6{\%} & \textbf{0.398}  & \textbf{0.0039}    \\
    \midrule
    GVB  &\multirow{3}{*}{\textit{\ourdata{}}} & 75.7\% & 68.3\% & 0.724 & 0.0095 \\
    RigNet & & 72.4\% & 71.1\%& 0.698  & 0.0091  \\
    Ours   &   & \textbf{80.7{\%}} & \textbf{77.2{\%}} & \textbf{0.337}  & \textbf{0.0050}    \\
    \bottomrule
  \end{tabular}
\end{table*}

\begin{table*}[]
  \caption{\textbf{Ablation studies on \res{} for skinning weight prediction.}}
  \label{ab_skin_supp}
  \centering
  \begin{tabular}{ccccc}
    \toprule
      &  Precision  &Recall & avg L1 & avg Dist. \\
    \midrule
    w/o geodesic dist.  & 81.5\% & 77.7\% & 0.444 & 0.0046 \\
    w/o weights norm   & 82.0\% & 77.9\%  & 0.436 & 0.0045 \\
     w/o shape features   & 81.4\% &
81.3\% &
0.412 & 0.0042 \\
    Ours      & \textbf{82.1{\%}} & \textbf{81.6{\%}} & \textbf{0.398}  & \textbf{0.0039}
    \\
    \bottomrule
  \end{tabular}
\end{table*}

\begin{table*}[h]
    \centering
    \caption{\textbf{Object counts for each category in the \ourdata{} dataset.}}
    \label{num_obj}
    \begin{tabular}{cccccc}
    \toprule
    Category        & \# Objects & Category        & \# Objects & Category        & \# Objects   \\
    \midrule
    character       & 16020      & miscellaneous & 584  & architecture & 132 \\
    anthropomorphic & 13393      &  scanned data &  546 & planet & 49 \\
    animal          & 4760       & plant  & 382 & paper & 46
    \\
    mythical creature & 4734      &  accessories & 293  &  musical instrument & 25
    \\
    toy         & 1360      & vehicle & 283 &  sporting goods & 21
    \\
    weapon          & 1257       & sculpture & 276 & armor & 13
    \\
    anatomy          & 1227      & household items & 274 & robot & 4
    \\
    clothing          & 595       & food &  206 \\
    \bottomrule
    \end{tabular}
\end{table*}

We provide additional qualitative comparisons among \ours{}, RigNet \cite{xu2020rignet}, and Pinocchio \cite{baran2007automatic} for skeleton generation.

\boldstartspace{More qualitative results on out-of-domain data.} 
We evaluate our method's generalization capability on diverse out-of-domain data sources: AI-generated meshes from Tripo2.0 \cite{tripo3d}, unregistered 3D scans from FAUST \cite{bogo2014faust}, and video-based 3D reconstructions \cite{song2024moda}.
As shown in \Cref{supp_skel_ood}, while existing methods struggle with generalization (RigNet fails across all cases, and Pinocchio shows misalignments even for human bodies, see skeleton results on the 3D scan), our method maintains robust performance across different data sources and categories. Notably, for human models, our method generates more detailed skeletal structures, including accurate hand skeletons, surpassing Pinocchio's template-based results.

\boldstartspace{More qualitative results on \ourdata{} and \res{}.}
We provide additional qualitative results on both \ourdata{} and \res{}  datasets. As illustrated in \Cref{supp_skel}, our method consistently generates high-quality skeletons that accurately match artist-created references across diverse object categories.

\begin{figure}
    \centering
\includegraphics[scale=0.35]{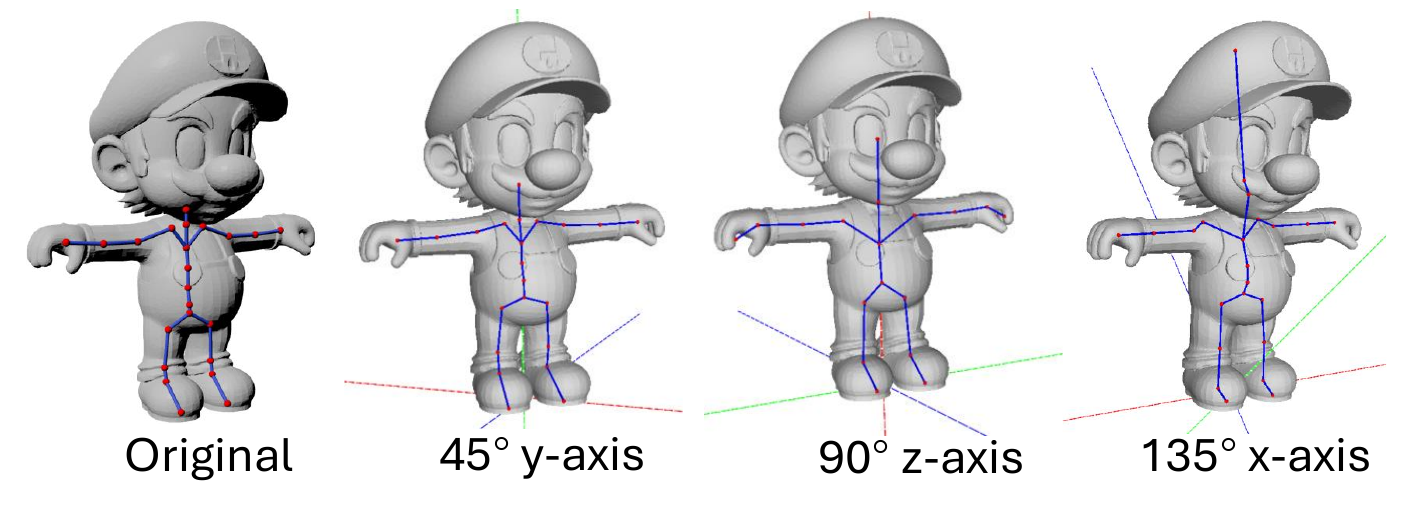}

    \caption{\textbf{Skeleton results on 3D models with different orientations.} Although minor differences may appear in the generated skeletons, all results maintain anatomically valid and suitable for rigging purposes.}
    \label{supp_rotation}
  \end{figure}

\boldstartspace{Robustness to various mesh orientations.}  To further validate our model’s robustness to various orientations, we include mesh rotations at multiple angles in \Cref{supp_rotation}. These examples show that our approach remains largely rotation-stable. While minor skeleton variations may occur, all generated results maintain anatomically valid and suitable for rigging purposes.

\subsection{More results of skinning weight prediction}
\boldstartspace{Quantitative results with deformation error.}
Beyond the precision, recall, and L1-norm metrics reported in the main paper, we evaluate the practical effectiveness of predicted skinning weights through deformation error analysis. This metric computes the average Euclidean distance between vertices deformed using predicted weights and ground truth weights across 10 random poses. The comprehensive results, shown in \Cref{comparison_skin_supp}, demonstrate our method's superior performance across most metrics on both datasets. We also include deformation error analysis in our ablation studies (\Cref{ab_skin_supp}), further validating the effectiveness of our design choices.

\begin{figure*}
    \centering
\includegraphics[width=\textwidth]
    {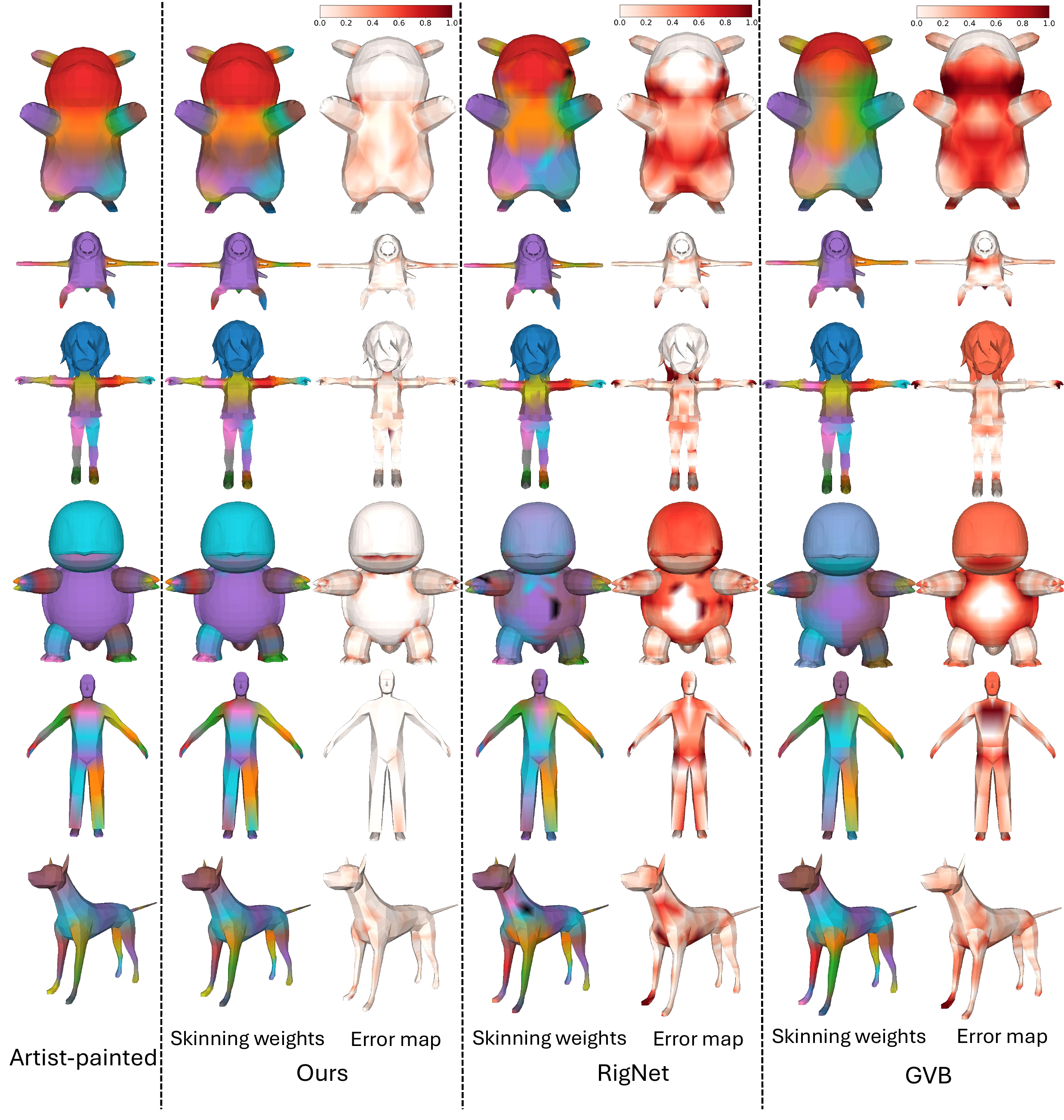}

    \caption{\textbf{Comparison of skinning weight prediction methods on \res{} (first three rows) and \ourdata{} (last three rows).} We visualize the predicted skinning weights alongside their corresponding L1 error maps.}
    \label{supp_skin}
  \end{figure*}

\boldstartspace{More qualitative results.}
We present additional qualitative comparisons between \ours{}, RigNet \cite{xu2020rignet}, and Geodesic Voxel Binding (GVB) \cite{dionne2013geodesic} for skinning weight prediction. \Cref{supp_skin} shows both the predicted skinning weights and their L1 error maps compared to artist-created references, demonstrating our method's superior accuracy across diverse object categories.

\section{More details of \ourdata{}}
\label{detail_data}

\subsection{Data Curation}

Our dataset curation process filters out duplicates, objects with extreme joint/bone counts, and multi-component objects. A detailed category-wise object distribution is provided in \Cref{num_obj}.

\subsection{Quality assessment}
\begin{figure*}
    \centering
    \includegraphics[scale=0.44]
    {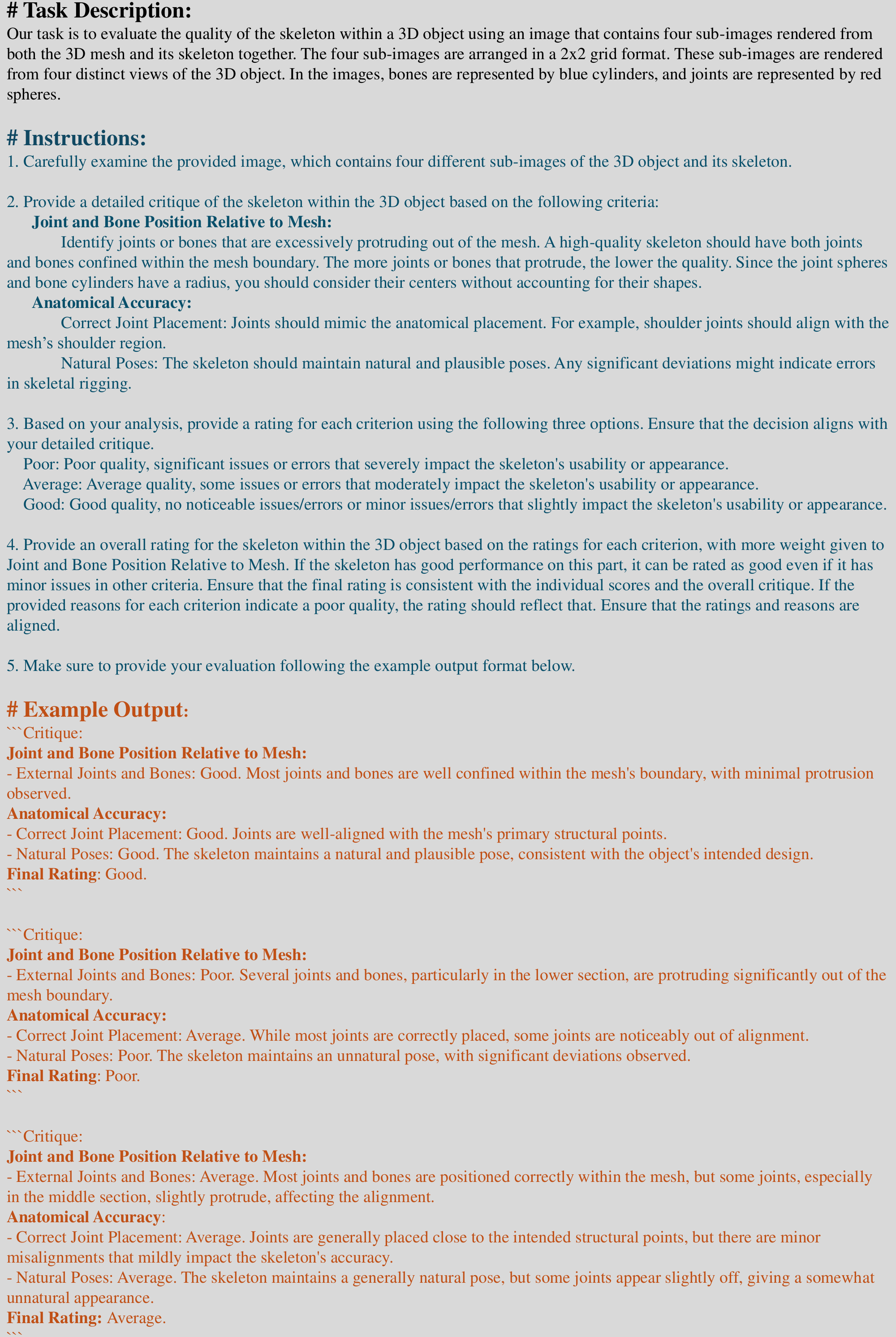}

    \caption{\textbf{Input instructions to VLM for data filtering.} }
    \label{rating_instruct}
  \end{figure*}

\begin{figure}
    \centering
    \includegraphics[scale=0.4]
{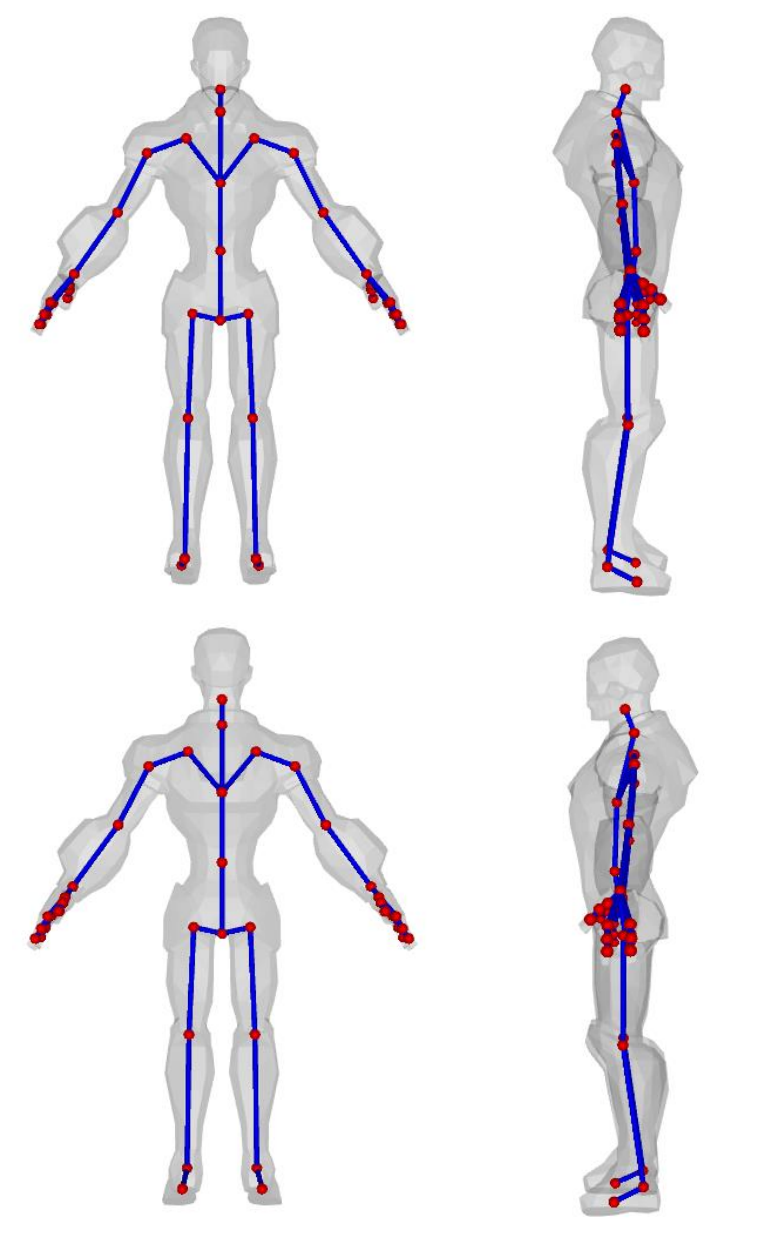}
    \caption{\textbf{Input rendered examples to VLM for data filtering.} }
    \label{rating_redner}
  \end{figure}

We employ GPT-4o \cite{openai_gpt4o} for quality assessment of skeleton annotations. 
For each model, we generate four-view renders using Pyrender\footnote{\url{https://github.com/mmatl/pyrender}} showing both the 3D model and its skeleton (\Cref{rating_redner}). These renders are evaluated using specific quality criteria detailed in \Cref{rating_instruct}.

\subsection{Category annotation}

\begin{figure*}
    \centering

\includegraphics[scale=0.48]
{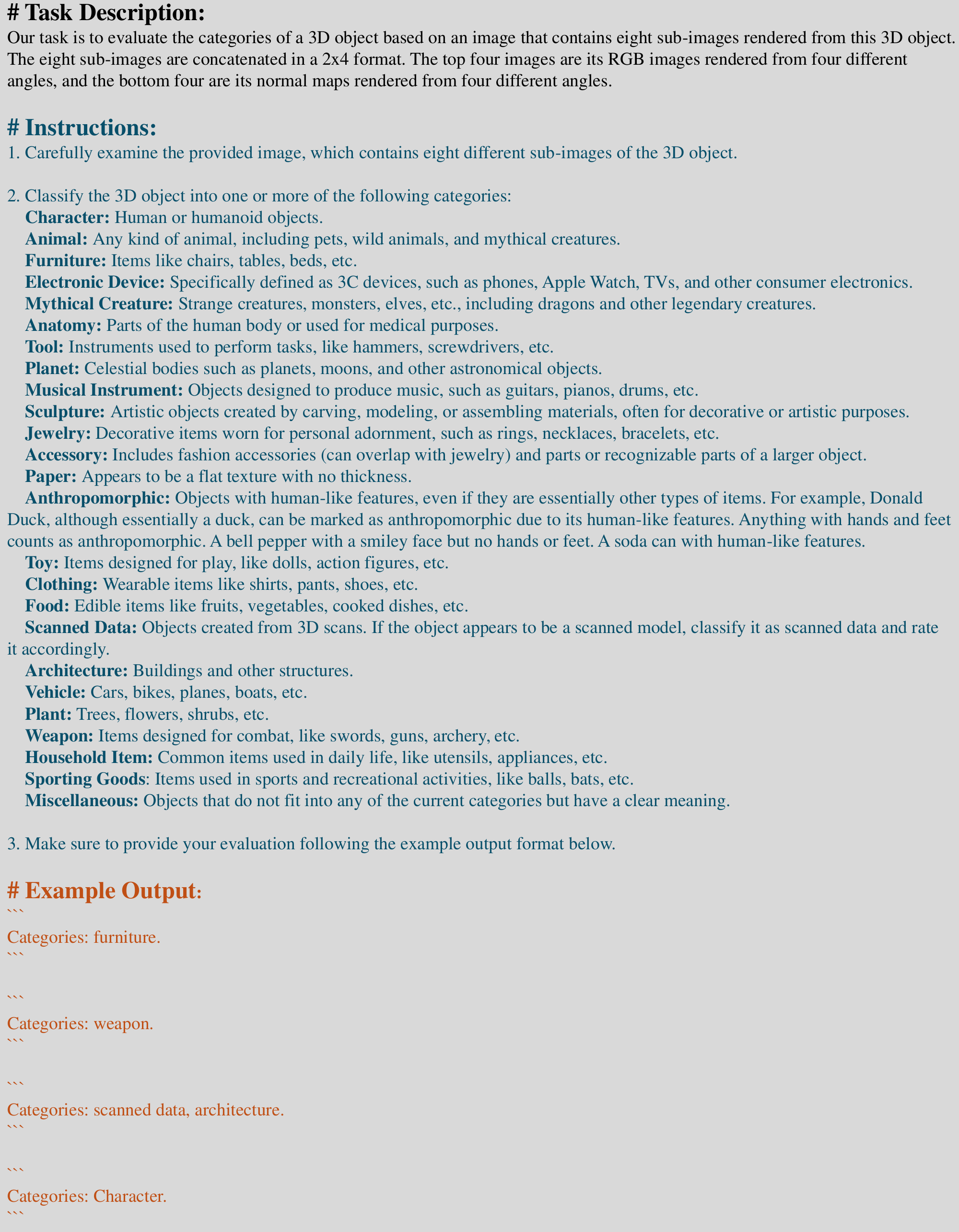}
    \caption{\textbf{Input instructions to VLM for category labeling.} }
    \label{cate_label_instruct}
  \end{figure*}

\begin{figure}
    \centering
   
    \includegraphics[scale=0.4]
{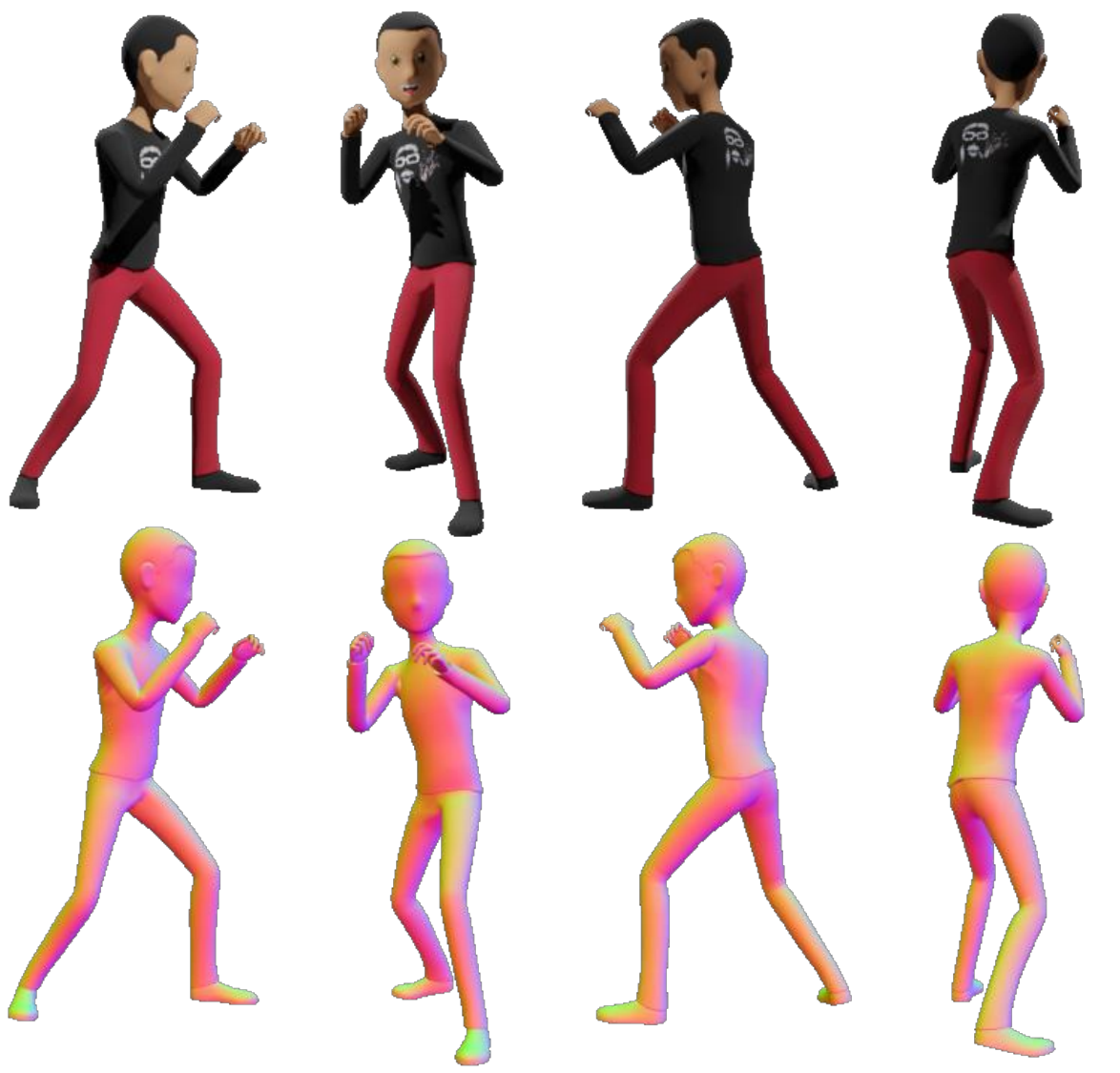}
    \caption{\textbf{Input rendered examples to VLM for category labeling.} }
    \label{cate_render}
  \end{figure}

For the Visual-Language Model (VLM)-based category labeling, we render each 3D model along with its normal maps from four viewpoints using Blender \cite{Blender} (see example in \Cref{cate_render}). We then utilize GPT-4o \cite{openai_gpt4o} to classify the categories of the 3D models based on specific instructions, as outlined in \Cref{cate_label_instruct}.

\section{Limitations and future work}
\label{limit}
\begin{figure}
    \centering
    \includegraphics[scale=0.14]
{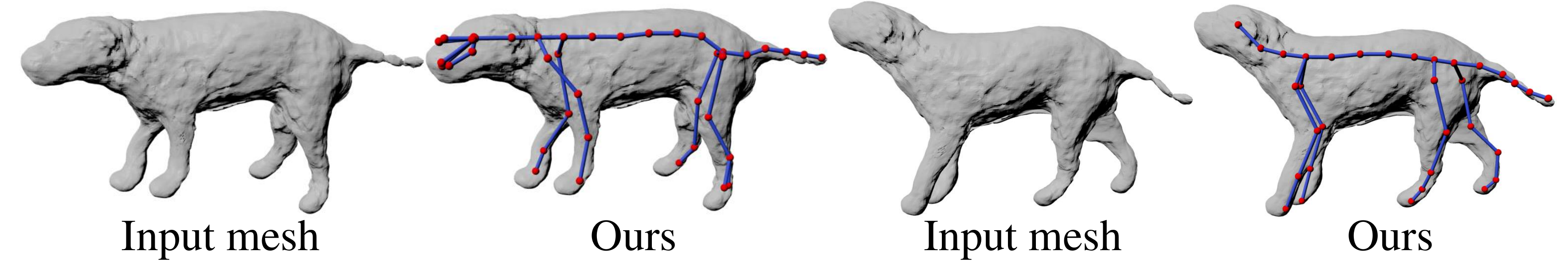}
    \caption{\textbf{Failure cases.} When input meshes possess very coarse surfaces (3D reconstruction results from \cite{song2023total}), our generated skeleton may exhibit inaccuracies, such as imperfect connections between the dog's trunk and legs.}
    \label{failure}
  \end{figure}

Despite its strong performance, our method has several notable limitations. First, our approach struggles with coarse mesh inputs, often producing inaccurate skeletons as shown in \Cref{failure}. While we employ preprocessing techniques to handle inputs from different sources, the significant domain gap between training data and coarse meshes remains challenging. Potential solutions include incorporating mesh quality augmentation during training to enhance robustness.

A second limitation lies in our dataset composition. Although \ourdata{} is large in scale, it lacks sufficient coverage of common articulated objects like laptops, staplers, and scissors, which affects our model's generalization to these categories.

Future work will address these limitations by:
1) Developing more robust preprocessing and training strategies for handling varying mesh qualities;
2) Expanding dataset coverage to include a broader range of everyday articulated objects;
3) Exploring techniques to better bridge the domain gap between different data sources.

\clearpage
{
    \small
    \bibliographystyle{ieeenat_fullname}
    \bibliography{main}
}

\end{document}